\pdfoutput=1

\documentclass[11pt]{article}

\usepackage[preprint]{acl}

\usepackage{times}
\usepackage{latexsym}

\usepackage[T1]{fontenc}

\usepackage[utf8]{inputenc}

\usepackage{microtype}

\usepackage{inconsolata}

\usepackage{graphicx}

\usepackage[utf8]{inputenc} 
\usepackage[T1]{fontenc}    
\usepackage{hyperref}       
\usepackage{url}            
\usepackage{booktabs}       
\usepackage{amsfonts}       
\usepackage{nicefrac}       
\usepackage{microtype}      
\usepackage{xcolor}         

\usepackage{algorithm}
\usepackage{algpseudocode}

\usepackage{graphicx}
\usepackage{caption}

\usepackage{amsmath}
\usepackage{amssymb}
\usepackage{mathtools}
\usepackage{amsthm}
\usepackage{array}
\usepackage{adjustbox}

\usepackage{subfigure}
\usepackage{float}
\usepackage{tabularx} 
\usepackage{subcaption}
\usepackage{multirow}    
\usepackage{tabularray}
\usepackage{makecell}

\usepackage[capitalize,noabbrev]{cleveref}

\title{Representing LLMs in Prompt Semantic Task Space}

\author{
  Idan Kashani \and Avi Mendelson \and Yaniv Nemcovsky \\
  Technion - Israel Institute of Technology \\
  Department of Computer Science \\
  \texttt{idan-kashani@cs.technion.ac.il}
}

\begin{document}
\maketitle

\begin{abstract}
Large language models (LLMs) achieve impressive results over various tasks, and ever-expanding public repositories contain an abundance of pre-trained models. Therefore, identifying the best-performing LLM for a given task is a significant challenge. Previous works have suggested learning LLM representations to address this. However, these approaches present limited scalability and require costly retraining to encompass additional models and datasets. Moreover, the produced representation utilizes distinct spaces that cannot be easily interpreted. This work presents an efficient, training-free approach to representing LLMs as linear operators within the prompts' semantic task space, thus providing a highly interpretable representation of the models' application. Our method utilizes closed-form computation of geometrical properties and ensures exceptional scalability and real-time adaptability to dynamically expanding repositories. We demonstrate our approach on success prediction and model selection tasks, achieving competitive or state-of-the-art results with notable performance in out-of-sample scenarios. 
\end{abstract}

\section{Introduction}

LLMs have recently emerged with remarkable capabilities, revolutionizing multiple diverse fields, including medical information processing ~\cite{zheng2024largelanguagemodelsmedicine,jin2024demystifyinglargelanguagemodels}, software engineering~\cite{etsenake2024understandinghumanllmdynamicliterature,jiang2024surveylargelanguagemodels,jimenez2024swebenchlanguagemodelsresolve}, and scientific research~\cite{frieder2024largelanguagemodelsmathematicians,zhang2024scientificlargelanguagemodels,li2024scilitllmadaptllmsscientific}. Moreover, open-source LLMs are widespread and have fueled a rapidly growing ecosystem of publicly-available models and benchmarks used to assess their capabilities. Currently, the most prominent platform hosting these resources is \textbf{Hugging Face}~\cite{wolf2019huggingface}, serving as a centralized repository for nearly one million pre-trained models and hundreds of thousands of benchmarks. Such open-access repositories 
facilitates LLMs' large-scale deployment and promotes their continuous innovation across diverse applications. 

The demand for LLM-based applications is ever-increasing, and new, diverse models with improved capabilities are constantly being produced. However, this rapid growth and diversity present a substantial challenge: \textit{identifying the best-performing models}. This challenge entails recognizing the most suitable model for producing a response to specific queries, or on average over queries in a given dataset. Hand-selecting these models would require careful analysis and clear annotation of their properties, which are not widely available. 

A common approach is to rely on benchmark results to select suitable models~\cite{chang2023surveyevaluationlargelanguage}. A \textbf{benchmark} consists of a dataset and an evaluation metric designed to assess specific model capabilities, such as domain expertise~\cite{hendrycks2021measuring,yu2024kola}, reasoning skills~\cite{parmar2023logicbench,pmlr-v162-velickovic22a,talmor-etal-2019-commonsenseqa}, agentic abilities~\cite{liu2024agentbench}, or safety~\cite{zhang-etal-2024-safetybench,li-etal-2024-salad,chao2024jailbreakbench}. While such benchmarking provides an initial assessment of models, it often involves a complex and time-consuming process to select a suitable model for a given prompt.
A primary challenge with conventional benchmarks lies in their reporting of aggregated performance scores, derived from a static corpus of domain-specific prompts. Such global metrics can be unreliable for guiding selection over queries that substantially diverge from the evaluation samples. Moreover, model proficiency often varies considerably across different prompts even within the same domain~\cite{zhuo-etal-2024-prosa,miller2024addingerrorbarsevals}. Typical benchmark outputs tend to obscure these instance-level performance nuances, thereby failing to provide the granular detail essential for precise, query-specific model selection. Furthermore, the perceived capabilities of LLMs can be skewed by inherent biases and sensitivities within benchmarks, resulting from particular prompt structures~\cite{cao2024on,pezeshkpour-hruschka-2024-large} or the methodologies behind leaderboard rankings~\cite{perlitz-etal-2024-efficient, alzahrani-etal-2024-benchmarks}. Beyond these limitations in granularity, relying on isolated benchmarks does not assess the collective insights available from the broader landscape of evaluation tools.

\textbf{Performance prediction} methods aim to predict models' performance on unseen prompts and tasks based on prior information. While such methods similarly utilize benchmarks, they differ in explicitly estimating models' performance over given queries, and aim to be applicable in scenarios where obtaining queries' labels is expensive or impractical. These approaches are particularly relevant to out-of-sample (OOS) settings, where the queries originate from datasets entirely unseen by the performance prediction estimators. Such settings introduce an additional layer of complexity, as generalization to unknown datasets is highly challenging. 

A recent work has suggested the approach of LLM embeddings for performance prediction and subsequent model selection~\cite{zhuang2025embedllm}. This approach aims to represent both LLMs and prompts in a joint space. The performance estimation is then computed via the corresponding embeddings of the model and query. However, current approaches utilize a distinct representation space that depends on their training data, thereby requiring costly retraining to include additional benchmark results. In this context, we denote the setting of real-time success prediction and subsequent model selection as aiming to apply to newly published models with minimal delay.

Our work builds on the promising direction of using LLM embeddings for performance prediction. We aim to represent LLMs as linear operators within the prompts' semantic task space, thus providing a highly interpretable representation of the models' application. We consider models' application on queries as semantic-space translations from input to output. We then utilize a closed-form computation to represent the difference between the model's induced and the desired translation, which produces the corresponding label. Our approach is training-free, requires negligible computational resources, and can be adapted to additional benchmarks in real-time. Moreover, we produce task-oriented embeddings with clear semantic interpretations relevant to diverse downstream tasks and OOS scenarios. Below we outline our main \textbf{contributions}:
\begin{enumerate}
    \item We present a novel approach to represent models directly within the semantic task of prompt embeddings. This direct representation presents a more intuitive and semantically grounded understanding of model-task relationships, enabling a more efficient and interpretable analysis of models' suitability.

    \item We utilize our representations for performance prediction, presenting an efficient and dynamically expandable evaluation of models. Our method utilizes closed-form computation, is training-free, and can be seamlessly expanded to additional models and benchmarks with negligible computational cost. Hereby, we enable real-time suitability analysis over the rapidly growing models-benchmarks ecosystem.

    \item We evaluate our method on performance prediction and model selection tasks over multiple settings and achieve state-of-the-art or comparable results. Moreover, our semantically grounded approach outperforms all previous baselines on OOS settings, indicating its robustness in diverse real-world scenarios.



\end{enumerate}

\section{Related Work}
\label{sec:related}

Previous LLM performance prediction works present various settings and approaches. Some works aim to analyze the behavior and performance of given models rather than directly predict their performance over given queries. A common setting discusses the assessment of model performance via per-sample output analysis, where approaches leverage confidence scores ~\cite{garg2022leveraging}, self-correction capabilities~\cite{jawahar-etal-2024-llm}, or transferability estimation~\cite{8803726, pmlr-v139-you21b,bassignana-etal-2022-evidence}. Another seeks to predict the broader capabilities of models through statistical analysis~\cite{4410411}, or by deriving scaling laws from pretraining data~\cite{chen2025scaling}. Although such approaches provide important perspectives, they are typically not applicable to the scope of this work.


Another set of methods trains auxiliary models to predict performance, for example, by training an assessor model which uses an LLM's results on fixed set of few reference prompts alongside target prompt intrinsic features to minimize evaluation costs~\cite{pacchiardi2025}, or by applying collaborative filtering to learn latent model and task factors from historical performance metadata~\cite{zhang-etal-2024-collaborative,drori2019automlusingmetadatalanguage,zhang2023automlgptautomaticmachinelearning}. These methods primarily contribute a trained predictive model designed to operate with specific inputs for the LLM in question. Our work, however, has a different underlying framework and is focused on deriving pre-computed, explicit vector representations for each LLM within an established library, using its comprehensive performance profile on source datasets.



The line of research most pertinent to our objective of creating explicit model representations from performance data involves \textbf{learning joint embeddings for LLMs and prompts}. EmbedLLM~\cite{zhuang2025embedllm} stands out as a key contribution in this domain. It employs an encoder-decoder architecture to learn informative representations from a large dataset of model-prompt interactions. While this work presents a significant step towards real-time performance prediction and subsequent model selection, it requires costly retraining to encompass additional models and benchmarks. Moreover, EmbedLLM's representations utilize an arbitrary space that lacks semantic grounding.

Our approach builds on the promising direction of using LLM embeddings for performance prediction, but aims to be dynamically expanding while representing LLMs within the prompts' semantic task space. 

\section{Method}
\label{sec:method}
\begin{figure}[tb]
 \centering
    \resizebox{\linewidth}{!}{
        \includegraphics[]{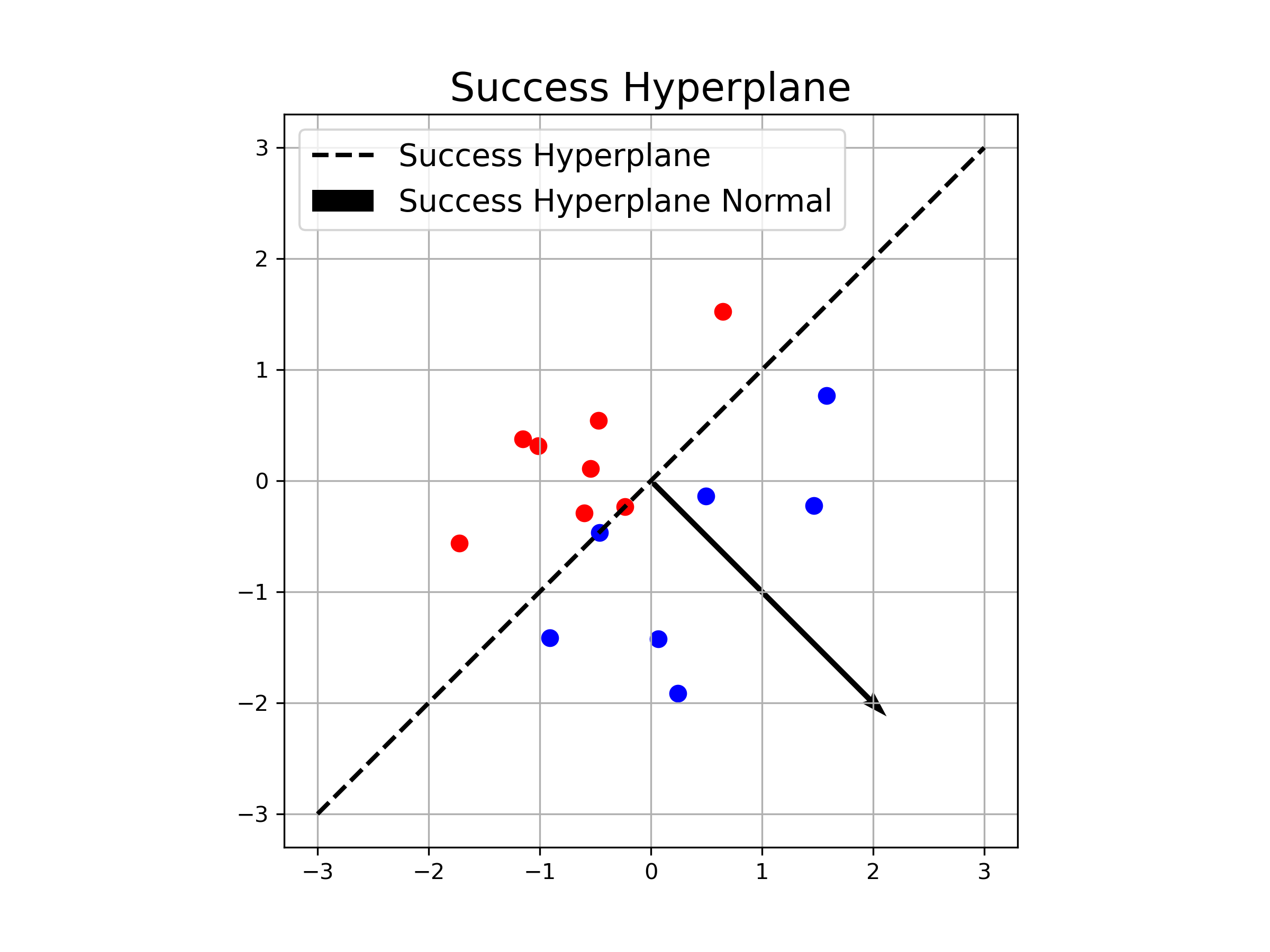}
    }
    \caption{
    The projection of a prompt embedding $E(p)$ on a model embedding $\mathbf{E(M)}_i$ yields a score predicting the model's success on that prompt.
    }
    \label{fig:succhyper}
\end{figure}

\begin{figure*}[t]
 \centering
    \resizebox{\linewidth}{!}{
        \begin{tabular}{cc}  
            \includegraphics[width=.48\textwidth]{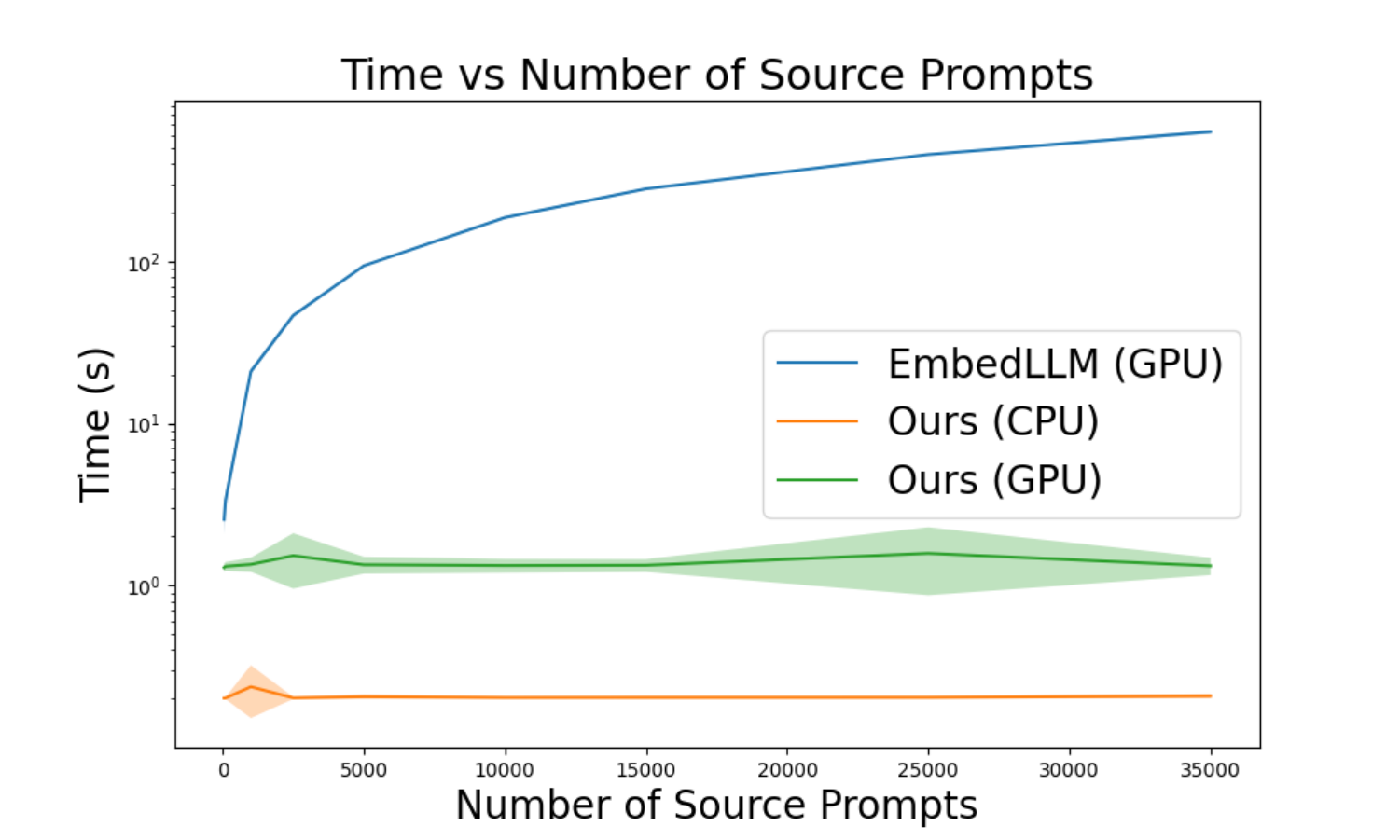}&
            \includegraphics[width=.48\textwidth]{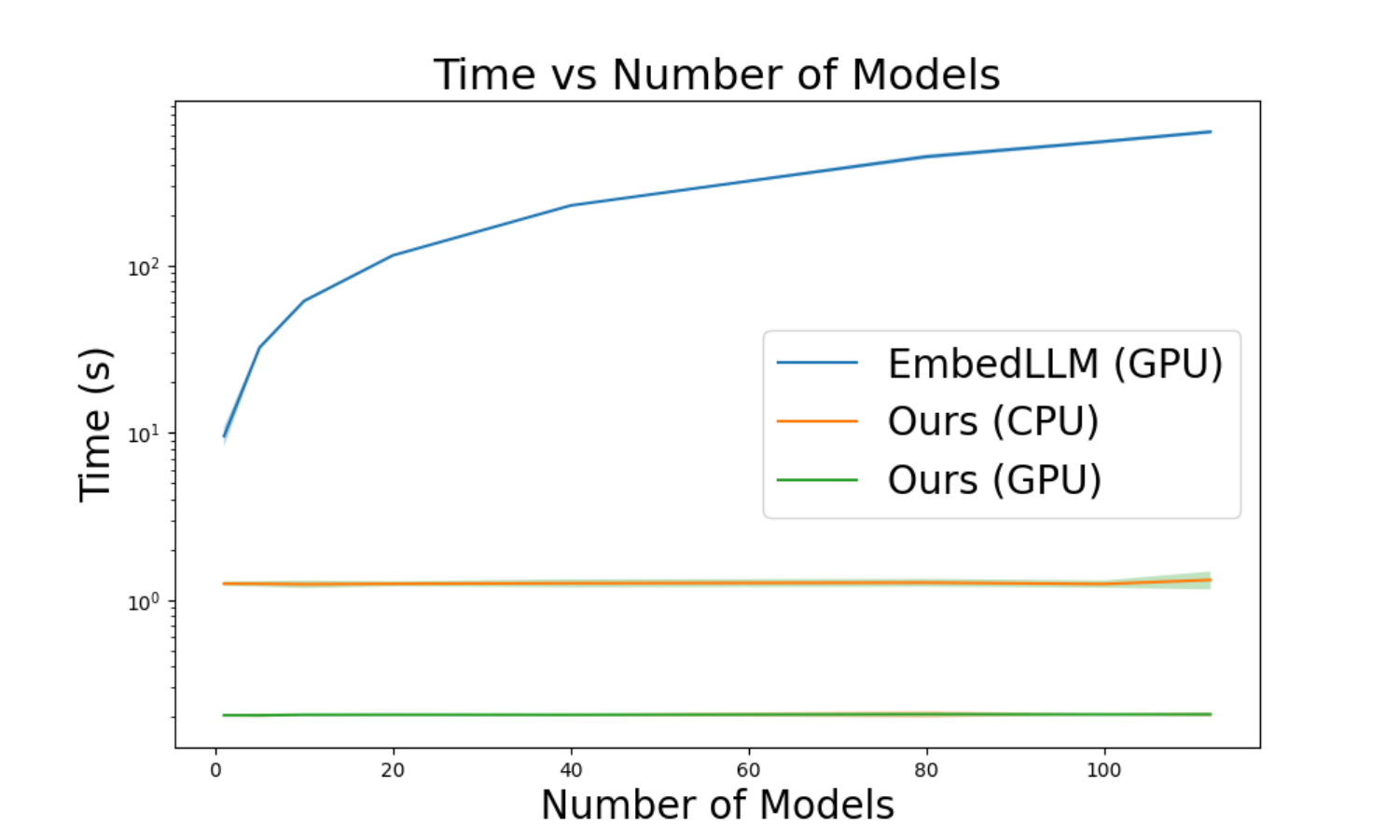}
        \end{tabular}
    }
    \caption{
    Model embeddings creation time vs.\@ number of prompt samples (left) and models (right), on CPU and GPU (logarithmic scale). 
    }
    \label{fig:emb_create_time}
\end{figure*}

We now detail our approach to creating linear, interpretable, and scalable representations for LLMs. We enable efficient performance prediction and model selection by embedding each LLM as a vector aligned with the prompts it successfully computes.

Formally, our goal is to derive a linear representation $\mathbf{E(M)}_i \in \mathbb{R}^{d_{\text{prompt}}}$ for each LLM $\mathcal{M}_i$ in a given pool $\mathcal{L}=\{\mathcal{M}_i\}_{i=1}^M$. This embedding $\mathbf{E(M)}_i$ is conceptualized as a vector in the $d_{\text{prompt}}$-dimensional prompt embedding space. Specifically, $\mathbf{E(M)}_i$ represents the ``\textbf{success hyperplane normal}'', namely a normal to a model-specific hyperplane that ideally separates between prompts where model $\mathcal{M}_i$ succeeds from those where it fails. The orientation of $\mathbf{E(M)}_i$ thus signifies a direction in the prompt space associated with success for that particular model. Consequently, the success of model $\mathcal{M}_i$ on a prompt $q$ is estimated by:
\begin{equation}
\hat{\text{Succ}}(\mathcal{M}_i, q) = \mathbf{E(M)}_i \cdot E(q). \label{eq:prediction_main}
\end{equation}

Here, $E(q)\in\mathbb{R}^{d_\text{prompt}}$ is the vector embedding for a given target prompt q, generated using the same pre-trained Sentence Transformer as the source prompts.


Conceptually, an LLM's success on a given prompt is a highly complex function of that prompt's semantic embedding, $f(E(q))$. Our method does not attempt to model $f$ in its entirety. Instead, we seek the best linear operator, represented by the vector $\mathbf{E(M)}_i$, that approximates the \textit{average} outcome of this function with respect to the success-failure dichotomy.
This approach is predicated on the well-established property of high-dimensional embedding spaces where semantic relationships can be represented as linear vector operations, a principle first established for word vectors~\cite{Mikolov2013EfficientEO} and since extended to produce robust sentence-level semantic representations~\cite{reimers-2019-sentence-bert}. The strong empirical success of our method suggests that this first-order linear approximation is sufficient to capture the most significant variance in model performance, offering a favorable trade-off between model fidelity and the exceptional scalability our approach provides.

Due to the linearity of our approach, the aggregate success score for a model on a benchmark can be efficiently computed by averaging the embeddings of the benchmark's prompts to form a single benchmark vector, then taking its dot product with the model's embedding.
\subsection{Data and System Formulation}
\label{sec:data_and_system}

To estimate these model embeddings $\mathbf{E(M)}$, we utilize:
\begin{itemize}
    \item A set of $N$ source prompts $p_j$ with corresponding ground-truth answers $a_j$, from $\mathcal{D_\text{src}}=\{(p_j, a_j)\}_{j=1}^N$.
    \item The observed performance of each of the $M$ LLMs from pool $\mathcal{L}$ on these prompts.
\end{itemize}

\paragraph{Prompt Embeddings.}
Each prompt $p_j$ is transformed into an $L^2$ normalized vector embedding $E(p_j)\in\mathbb{R}^{d_\text{prompt}}$ using a pre-trained \textbf{Sentence Transformer}. This type of architecture (e.g., based on~\cite{devlin-etal-2019-bert, Schroff_2015_CVPR}) is chosen for its efficiency and established ability to capture semantic content relevant for comparing text sequences. These prompt embeddings form the rows of a matrix $\mathbf{D_\textbf{src}}\in\mathbb{R}^{N \times d_\text{prompt}}$.

\paragraph{Performance Matrix.}
Benchmarks results measure a certain property of a model with respect to a dataset.
\textbf{Success} is determined using an \textbf{exact match} criterion between a model $\mathcal{M}_i$'s output for prompt $p_j$ and the target answer $a_j$. This binary outcome (success/failure) is encoded in a performance matrix $\mathbf{P_\textbf{src}} \in \mathbb{R}^{M \times N}$, where:
$$
\mathbf{P_{\textbf{src}_{ij}}} = \begin{cases}
             \hspace{2.78mm}1, & \text{if } \mathcal{M}_i(p_j) = a_j\\
            -1, & \text{otherwise.}
\end{cases}
$$

\paragraph{The Linear System.}
Given our conceptualization of model embeddings (\Cref{eq:prediction_main}), the relationship $\mathbf{E(M)}_i \cdot E(p_j) \approx \mathbf{P_{\textbf{src}_{ij}}}$ should hold for all models and prompts. This can be expressed in matrix form as the linear system we aim to solve for $\mathbf{E(M)}$:
\begin{equation}
\mathbf{E(M)} (\mathbf{D_\textbf{src}})^\intercal \approx \mathbf{P_\textbf{src}}.
\label{eq:linear_system_to_solve}
\end{equation}

\subsection{Computing Linear LLM Representations}
\label{sec:computing_embeddings}

We solve the linear system (\Cref{eq:linear_system_to_solve}) for $\mathbf{E(M)}$. Since the matrix $(\mathbf{D_\textbf{src}})^\intercal$ (derived from prompt embeddings) is typically non-square and may be non-invertible, we employ its regularized Moore-Penrose pseudoinverse~\cite{moore1920reciprocal, bjerhammar1951application, penrose1955generalized,ben2003generalized}, computed via Singular Value Decomposition (SVD)~\cite{svdeckart1936approximation}. The regularization is crucial for stability and to improve generalization to unseen data:

\begin{itemize}
    \item \textbf{Singular Value Thresholding:} Singular values ($\sigma_{ii}$) from the SVD of $\mathbf{D_\textbf{src}}$ that fall below a predefined threshold, $\varepsilon$, are effectively set to zero before forming the pseudoinverse components\footnote{Not to be confused with the numerical tolerance threshold $t = \text{machine precision}\cdot\max(N, d_{\text{prompt}})\cdot\max(\text{diag}(\Sigma))$. In our experiments, $\varepsilon > t$.}. This mitigates numerical instabilities from near-zero singular values. We found the choice of this threshold ($\varepsilon$) significantly affects results, especially for OOS settings (\cref{sec:appendix_a}).
    \item \textbf{Tikhonov Regularization:} We apply Tikhonov regularization to smooth the inversion. This is achieved by incorporating the regularization term $2\lambda$ with the squared singular values when deriving the effective inverse singular values~\cite{tikhonov1943stability,hoerl1970ridge2,hoerl1970ridge1}.
\end{itemize}

The closed-form solution for the model embeddings $\mathbf{E(M)}$ is:
\begin{equation}
    \mathbf{E(M)} = \mathbf{P_\textbf{src}}(\mathbf{D_\textbf{src}^+})^\intercal=\mathbf{P_\textbf{src}}\mathbf{U\Sigma'V^\intercal}, \label{eq:final_EM_solution}
\end{equation}
where $\mathbf{D_\textbf{src}}=\mathbf{U \Sigma V^\intercal}$ is the SVD of the prompt embedding matrix. The diagonal matrix $\mathbf{\Sigma'}$ is derived from the singular values $\sigma_{ii}$ in $\mathbf{\Sigma}$, with its elements are given by:
\begin{align} 
\sigma'_{ii} &= \begin{cases} 
    0, & \text{if } \sigma_{ii} < \varepsilon\\
    \frac{\sigma_{ii}}{ \sigma_{ii}^2 + 2\lambda}, & \text{otherwise}
\end{cases} \label{eq:sigma_prime_computation} \\ 
\sigma'_{ij} &= 0, \qquad \text{for} \quad i \neq j. \nonumber
\end{align}
The resulting model embedding matrix $\mathbf{E(M)}$ is of size $M\times d_{\text{prompt}}$.
This linear, training-free computation makes our approach highly efficient and directly interpretable within the prompt's semantic space.

\subsection{Scalability Analysis}
\label{sec:scalability_analysis}

The primary computational cost is the SVD of $\mathbf{D_\textbf{src}}$ (dimensions $N \times d_{\text{prompt}}$), which has a complexity of $O(Nd_{\text{prompt}}^2)$ and is thus linear in $N$ for a fixed $d_{\text{prompt}}$. The subsequent matrix multiplication to obtain $\mathbf{E(M)}$ is linear in both $M$ and $N$. This results in significantly better overall scalability compared to EmbedLLM, whose training time typically exhibits much steeper growth with increasing $M$ or $N$. Further efficiency can be achieved by applying iterative methods to update $(\mathbf{D_\textbf{src}^+})^\intercal$, such as the Newton-Schulz iteration (\Cref{sec:appendix_incremental}).

The method also offers theoretical stability for distributed systems. Since the embeddings are based on the semantic space of prompts, adding new prompts to a dataset is expected to induce only minor changes to existing model embeddings. This leads to minimal discrepancies in the embeddings over time across different servers.

\section{Experiments}
\label{sec:exp}


This section presents a comprehensive empirical evaluation of our method over the tasks of real-time \textbf{success prediction} and subsequent \textbf{model selection}. We first present the experimental setting in \cref{sec:exp_setup}, and continue to discuss the results in \cref{sec:exp_results}. In addition, we present an ablation study of our method in supplementary \cref{sec:appendix_a}. Our evaluation aims to answer three key research questions:

\begin{itemize}
    \item \textbf{(RQ1)} How does our method compare to existing approaches, regarding predicted success and model selection?
    \item \textbf{(RQ2)} How does the scalability of our proposed method compare to previous work?
    \item \textbf{(RQ3)} Does our method present a viable approach to predict success in OOS settings?
\end{itemize}



\subsection{Experimental Setup}
\label{sec:exp_setup}

    \subsubsection{Core Evaluation Tasks and Metrics}
    \label{sec:exp_tasks_metrics}
    \begin{figure*}[tb]
 \centering
    \resizebox{.96\textwidth}{!}{
        \begin{tabular}{cc}  
            \includegraphics[width=.48\textwidth]{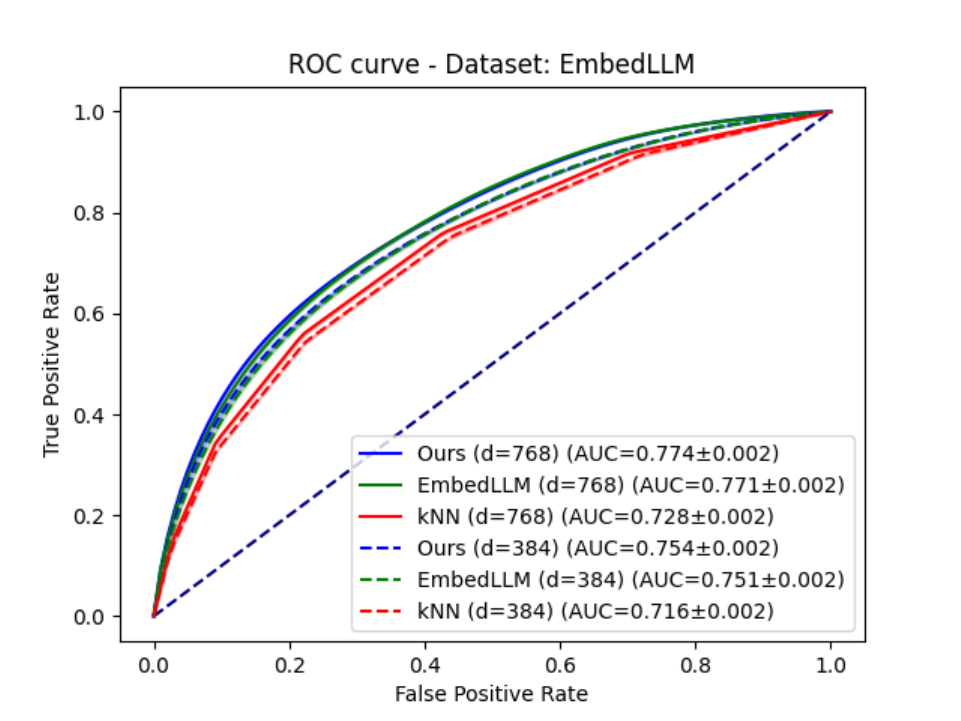} &
            \includegraphics[width=.48\textwidth]{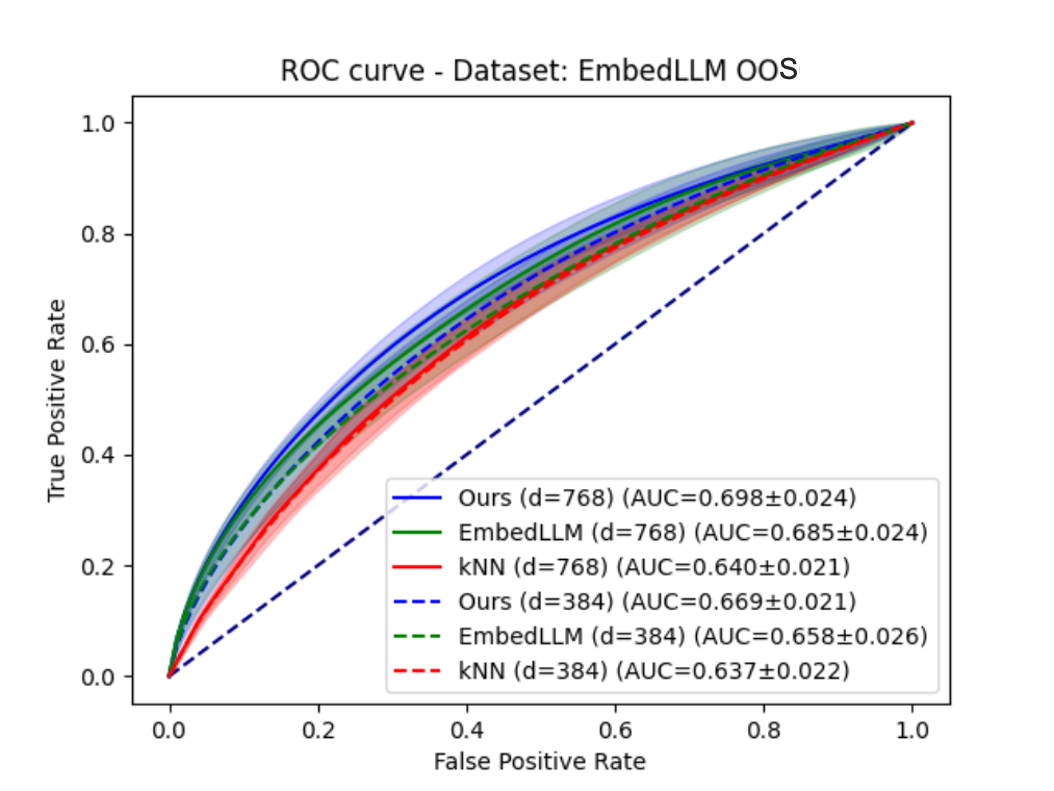} \\
            \includegraphics[width=.48\textwidth]{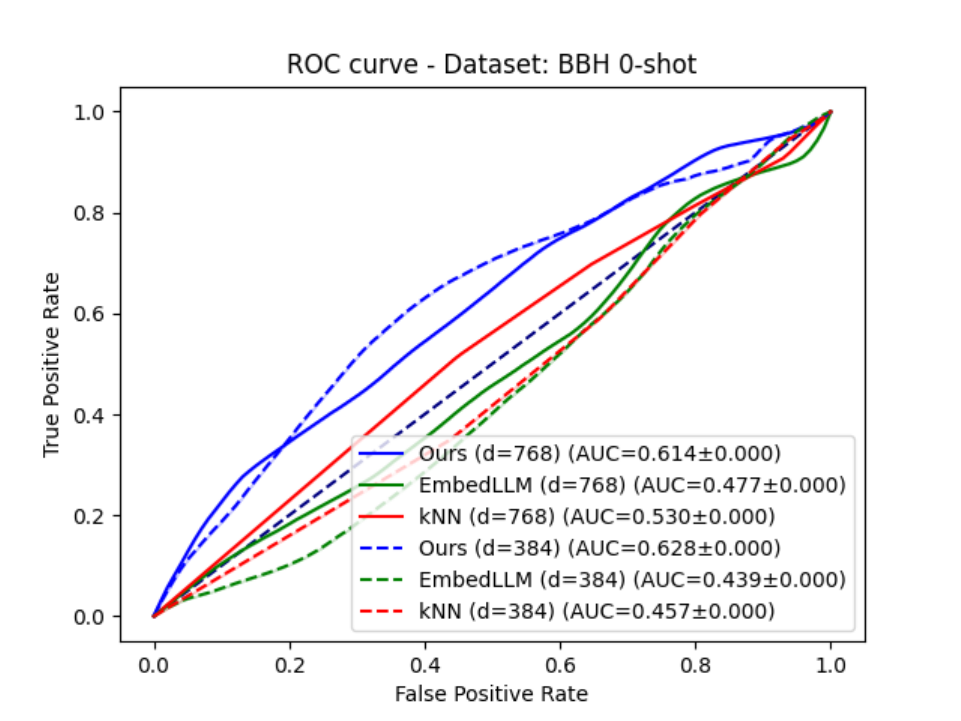} &
            \includegraphics[width=.48\textwidth]{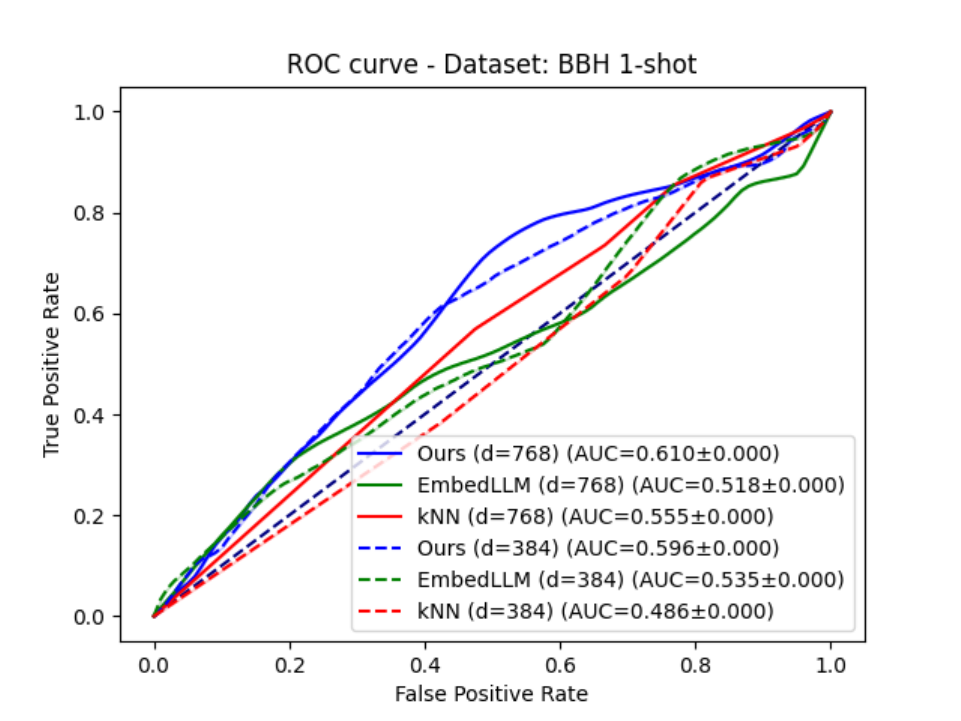}
        \end{tabular}
    }
    \caption{
    Success Prediction ROC curves describe the true positive rate vs. the false positive rate across thresholds.
    }
    \label{fig:pp_roc_curves}
\end{figure*}

    \paragraph{Success Prediction.} This task presents a binary classification problem, i.e., given an LLM and a prompt, will the model successfully complete the task defined by the prompt? The task performance metrics are then:
    \begin{itemize}
        \item \textbf{AUC (Area Under the ROC Curve):} Measures the ability to distinguish between success and failure, irrespective of a specific classification threshold.
        \item \textbf{Accuracy:} The fraction of correct success/failure predictions.
        \item \textbf{Benchmark Score Correlation:} The Pearson correlation between our method's estimated success scores for a model across a benchmark's prompts (which can be computed efficiently as presented in \Cref{sec:method}) and the model's actual ground-truth accuracy on that benchmark.
    \end{itemize}

    \paragraph{Model Selection.} In this task, given sets of models and test prompts, methods aim to accurately rank the models according to their expected success over the prompts. The task performance metrics are then:
    \begin{itemize}
        \item \textbf{Accuracy:} The proportion of test prompts for which the best-ranked (selected) model produces a successful response.
        \item \textbf{Recall:} The proportion of ``solvable'' prompts over which the top-ranked model succeeds. This metric evaluates the selector's ability to choose a successful model for prompts that \textit{at least one} model in the pool can solve. It directly measures how the selected model's performance compares to the best possible outcome on a per-prompt basis
    \end{itemize}

    \subsubsection{Models \& Datasets Environments}
    \label{sec:exp_datasets_env}
    Our experiments were conducted on four distinct environments, each comprising specific sets of models, source prompts used for methods' execution, and target prompts used for evaluation. These are derived from two primary sources:
    \begin{enumerate}
        \item \textbf{EmbedLLM Benchmark Environment}:
        \begin{itemize}
            \item \textbf{Models}: 112 LLMs from the EmbedLLM framework, covering both general-purpose and domain-specific architectures.
            \item \textbf{Source \& Target prompts}: Source and target prompts are randomly sampled in either in-sample or OOS scenarios from over $80$ prominent benchmarks, such as MathQA~\cite{amini-etal-2019-mathqa}, SocialQA~\cite{sap-etal-2019-social}, PIQA~\cite{piqa}, LogiQA~\cite{ijcai2020p501}, ASDiv~\cite{miao-etal-2020-diverse}, GSM8K~\cite{cobbe2021trainingverifierssolvemath}, MMLU~\cite{hendrycks2021measuring}, TruthfulQA~\cite{Lin2021TruthfulQAMH}, MedMCQA~\cite{pmlr-v174-pal22a}, and GPQA~\cite{rein2024gpqa}. For the EmbedLLM environment, both in-sample and OOS results are the average of 10 independent trials. In each trial, a new random seed is used to sample the source and target prompts from the available benchmarks. The error margins in \Cref{tab:sp_ms} and the shaded regions in the ROC curves of \Cref{fig:pp_roc_curves} represent the standard deviation across these 10 trials.
            \item \textbf{In-Sample Scenario}: Test prompts and source prompts are sampled may originate from the same datasets.
            \item \textbf{Out-of-Sample (OOS) Scenario}: Test datasets are excluded from the datasets library, that is used as source prompts for the method. This scenario represents a more complex generalization task.
        \end{itemize}
        \item \textbf{LoRA-Finetuned T5 Models Environment}: 
        \begin{itemize}
            \item \textbf{Models}: A set of 92 T5-large FLAN encoder-decoder models. These models were finetuned on FLAN v2 datasets~\cite{JMLR:v25:23-0870} using the LoRA technique~\cite{hu2022lora} by \citet{huang2024lorahub}.
            \item \textbf{Source Datasets}: The 92 FLAN v2 datasets used for the original LoRA training (listed in \Cref{sec:appendix_b}).
            \item \textbf{Target Datasets (OOS)}: BIG-Bench Hard (BBH)~\cite{suzgun-etal-2023-challenging}, a suite of 23 challenging tasks for LLMs.
            \item \textbf{Evaluation Scenarios}: Performance is evaluated in either OOS zero-shot or OOS one-shot settings. One demonstration prompt is sampled per dataset for the one-shot.
        \end{itemize}
    \end{enumerate}

    \subsubsection{Baselines and Method Configuration}
    \label{sec:exp_baselines_config}

    \paragraph{Prompt Embeddings.} We utilize two pre-trained Sentence Transformer models to generate $L_2$ normalized prompt embeddings:
    \begin{itemize}
        \item MiniLM-L6-v2 ($d_{\text{prompt}}=384$)~\cite{wang-etal-2021-minilmv2}\footnote{\url{https://huggingface.co/sentence-transformers/all-MiniLM-L6-v2}}
        \item MPNet-v2 ($d_{\text{prompt}}=768$)~\cite{NEURIPS2020_c3a690be}\footnote{\url{https://huggingface.co/sentence-transformers/all-mpnet-base-v2}}
    \end{itemize}

    \paragraph{Previous art.}
    Our method is compared against two primary baselines, following the EmbedLLM paper~\cite{zhuang2025embedllm}:
    \begin{enumerate}
        \item \textbf{k-Nearest Neighbors (kNN):} Configured with $k=5$. For success prediction on a target prompt, kNN identifies the $k$ nearest prompts in the source data via prompt embedding similarity. The average success rate of the target model on these neighbors is then the predicted score. This evaluation metric is then utilized in model selection by selecting the candidate LLM with the maximal score.
        \item \textbf{EmbedLLM:} Evaluated using the provided framework.
    \end{enumerate}
    A comparison with an additional static baseline, ``Best Source Performer'', is provided in \Cref{sec:appendix_c}.

    \paragraph{Our Configuration.} We utilize the regularization parameter $\lambda=1$ in all the compared settings. The optimal singular value threshold $\varepsilon$ is then determined via the ablation study presented in \Cref{sec:appendix_a}.

\begin{table*}[tb] 
\centering
\caption{Comparison of \textit{success prediction} and \textit{model selection} across methods and prompt embedding dimension.} 
\label{tab:sp_ms}
\Large
\resizebox{\textwidth}{!}{
\setlength{\tabcolsep}{3pt} 
\begin{tabular}{@{}llcccccc@{}} 
\toprule\toprule
\multicolumn{3}{c}{} & \multicolumn{3}{c}{Success Prediction} & \multicolumn{2}{c}{Model Selection} \\
\cmidrule(lr){4-6} \cmidrule(lr){7-8} 
Environment & Method & dim & AUC & Accuracy & Benchmark Score Correlation & Accuracy & Recall \\
\midrule\midrule

\multirow{6}{*}{\textbf{EmbedLLM}} & kNN & \multirow{3}{*}{384} & $0.7158 \pm 0.0019$ & $0.6855 \pm 0.0009$ & $0.7665 \pm 0.0139$ & $0.5261 \pm 0.0029$ & $0.5762 \pm 0.0033$ \\
& EmbedLLM & & \underline{$0.7509 \pm 0.0018$} & \underline{$0.7076 \pm 0.0013$} & \underline{$0.9030 \pm 0.0022$} & $\mathbf{0.6269 \pm 0.0024}$ & $\mathbf{0.6867 \pm 0.0030}$ \\
& Ours & & $\mathbf{0.7538 \pm 0.0019}$ & $\mathbf{0.7115 \pm 0.0015}$ & $\mathbf{0.9248 \pm 0.0027}$ & $\underline{0.6221 \pm 0.0020}$ & $\underline{0.6814 \pm 0.0027}$ \\
\cmidrule(lr){2-8} 
& kNN & \multirow{3}{*}{768} & $0.7285 \pm 0.0019$ & $0.6937 \pm 0.0014$ & $0.7498 \pm 0.0123$ & $0.5335 \pm 0.0031$ & $0.5844 \pm 0.0033$ \\
& EmbedLLM & & \underline{$0.7714 \pm 0.0018$} & \underline{$0.7183 \pm 0.0014$} & \underline{$0.9266 \pm 0.0022$} & $\mathbf{0.6410 \pm 0.0018}$ & $\mathbf{0.7022 \pm 0.0025}$ \\
& Ours & & $\mathbf{0.7736 \pm 0.0018}$ & $\mathbf{0.7232 \pm 0.0013}$ & $\mathbf{0.9485 \pm 0.0025}$ & $\underline{0.6355 \pm 0.0012}$ & $\underline{0.6961 \pm 0.0014}$ \\
\midrule\midrule

\multirow{6}{*}{\textbf{EmbedLLM (OOS)}}
& kNN & \multirow{3}{*}{384} & $0.6366 \pm 0.0224$ & $0.6205 \pm 0.0262$ & $0.6971 \pm 0.0531$ & $0.4779 \pm 0.0426$ & $0.5198 \pm 0.0448$ \\
& EmbedLLM & & \underline{$0.6580 \pm 0.0259$} & \underline{$0.6466 \pm 0.0266$} & \underline{$0.8310 \pm 0.0323$} & $\underline{0.5667 \pm 0.0635}$ & $\underline{0.6165 \pm 0.0688}$ \\
& Ours & & $\mathbf{0.6696 \pm 0.0214}$ & $\mathbf{0.6480 \pm 0.0242}$ & $\mathbf{0.8451 \pm 0.0300}$ & $\mathbf{0.5879 \pm 0.0441}$ & $\mathbf{0.6393 \pm 0.0413}$ \\
\cmidrule(lr){2-8}
& kNN & \multirow{3}{*}{768} & $0.6404 \pm 0.0211$ & $0.6230 \pm 0.0216$ & $0.6860 \pm 0.0311$ & $0.4807 \pm 0.0365$ & $0.5229 \pm 0.0372$ \\
& EmbedLLM & & \underline{$0.6848 \pm 0.0239$} & \underline{$0.6622 \pm 0.0247$} & \underline{$0.8663 \pm 0.0292$} & $\underline{0.5694 \pm 0.0640}$ & $\underline{0.6194 \pm 0.0682}$ \\
& Ours & & $\mathbf{0.6983 \pm 0.0240}$ & $\mathbf{0.6694 \pm 0.0153}$ & $\mathbf{0.8820 \pm 0.0322}$ & $\mathbf{0.5916 \pm 0.0436}$ & $\mathbf{0.6435 \pm 0.0435}$ \\
\midrule\midrule

\multirow{6}{*}{\textbf{BBH 0-shot (OOS)}}
& kNN & \multirow{3}{*}{384} & \underline{$0.4573$} & $0.3251$ & $-0.2376$ & $0.2014$ & $0.4766$ \\
& EmbedLLM & & $0.4394$ & \underline{$0.3297$} & \underline{$-0.1408$} & $\underline{0.2275}$ & $\underline{0.5383}$ \\
& Ours & & $\mathbf{0.6284}$ & $\mathbf{0.4351}$ & $\mathbf{0.3734}$ & $\mathbf{0.2491}$ & $\mathbf{0.5896}$ \\
\cmidrule(lr){2-8}
& kNN & \multirow{3}{*}{768} & \underline{$0.5301$} & \underline{$0.3351$} & \underline{$0.1353$} & $0.1971$ & $0.4664$ \\
& EmbedLLM & & $0.4769$ & $0.2937$ & $-0.0766$ & $\underline{0.2113}$ & $\underline{0.5002}$ \\
& Ours & & $\mathbf{0.6139}$ & $\mathbf{0.3838}$ & $\mathbf{0.3862}$ & $\mathbf{0.2491}$ & $\mathbf{0.5896}$ \\
\midrule\midrule

\multirow{6}{*}{\textbf{BBH 1-shot (OOS)}} & kNN & \multirow{3}{*}{384} & $0.4858$ & $0.4132$ & $-0.0528$ & $0.3330$ & $0.7012$ \\
& EmbedLLM & & \underline{$0.5353$} & \underline{$0.4347$} & \underline{$0.1116$} & $\mathbf{0.3408}$ & $\mathbf{0.7178}$ \\
& Ours & & $\mathbf{0.6546}$ & $\mathbf{0.5034}$ & $\mathbf{0.5569}$ & $\underline{0.3405}$ & $\underline{0.7171}$ \\
\cmidrule(lr){2-8}
& kNN & \multirow{3}{*}{768} & \underline{$0.5550$} & \underline{$0.4390$} & \underline{$0.2757$} & $0.3251$ & $0.6846$ \\
& EmbedLLM & & $0.5178$ & $0.3298$ & $-0.0562$ & $\underline{0.3274}$ & $\underline{0.6895}$ \\
& Ours & & $\mathbf{0.6097}$ & $\mathbf{0.4631}$ & $\mathbf{0.3843}$ & $\mathbf{0.3381}$ & $\mathbf{0.7119}$ \\
\bottomrule\bottomrule
\end{tabular}
} 
\end{table*}
\begin{figure*}[t]
\centering

            

\includegraphics[width=0.97\linewidth]{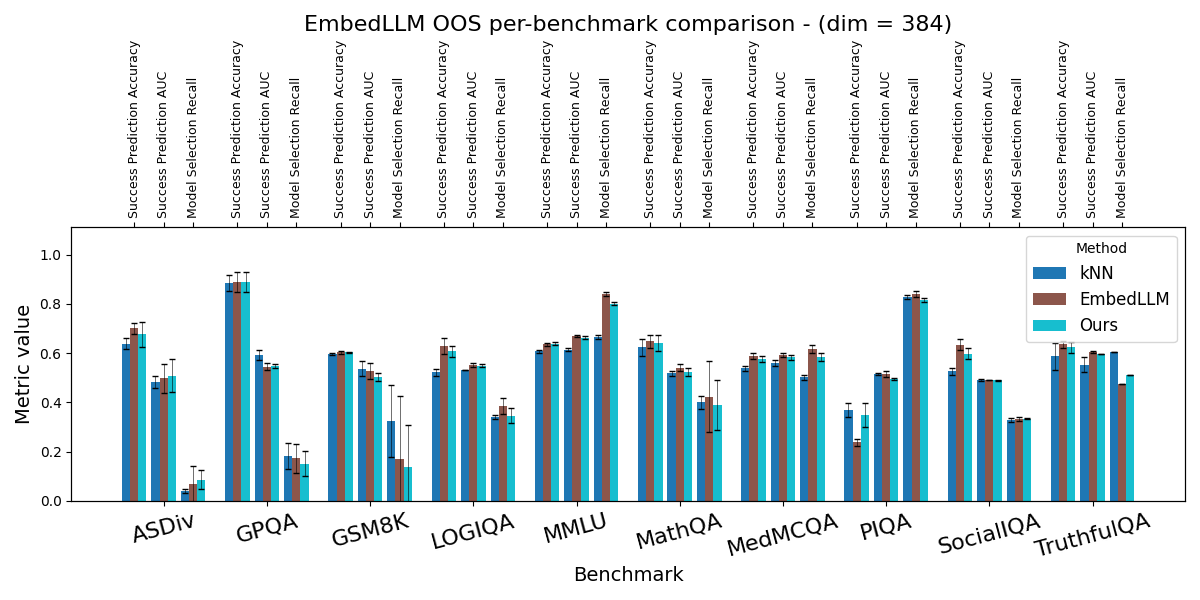}
\vspace{0.4em}
\includegraphics[width=0.97\linewidth]{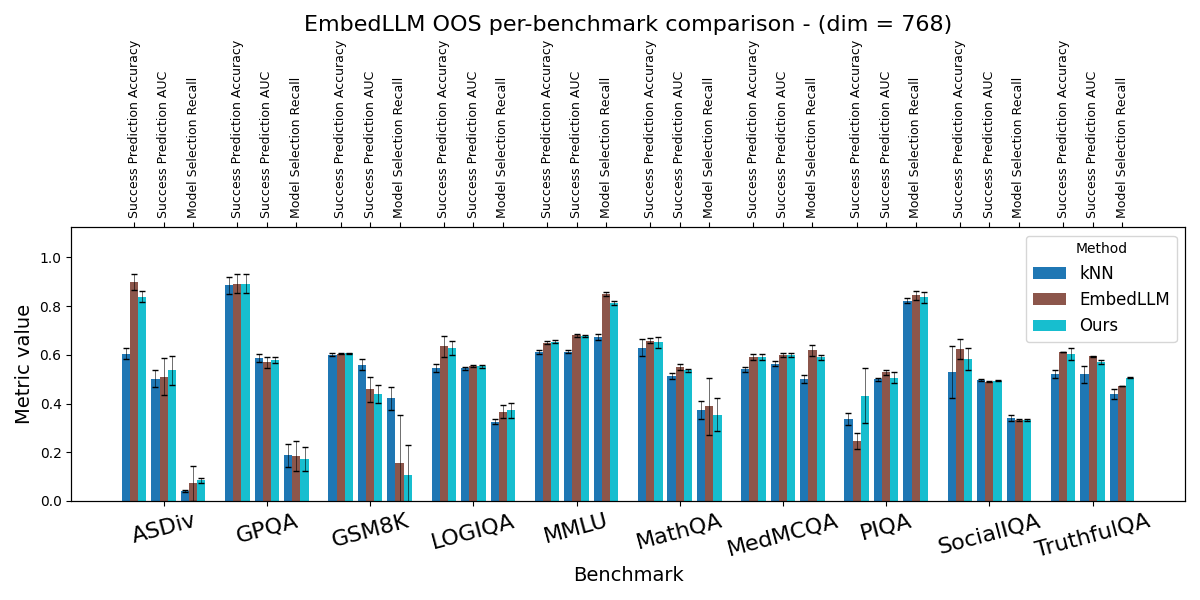}
\caption{A per-benchmark breakdown of \textit{Success Prediction (Accuracy and AUC)} and \textit{Model Selection (Recall)} in the EmbedLLM OOS environment for embedding dimensions 384 (top) and 768 (bottom). Our training-free method delivers performance competitive with the EmbedLLM baseline at a fraction of the computational cost.
}
\label{fig:per_benchmark_comparison}
\end{figure*}

\subsection{Experimental Results}
\label{sec:exp_results}


    \subsubsection{Success Prediction}
    
    In \Cref{tab:sp_ms} and \Cref{fig:pp_roc_curves}, we present the comparison of our method to previous art over the success prediction task. Our method achieves a better ratio between true positives and false positives compared to previous art. Moreover, it outperforms all previous art over AUC, Accuracy, and Benchmark Score Correlation, substantially so in the OOS scenarios (EmbedLLM (OOS), BBH 0-shot, and BBH 1-shot). 
    This may suggest that our semantically grounded approach effectively captures fundamental model-task alignment and is better suited for generalization to unseen datasets. The high benchmark score correlation further indicates that our lightweight, per-prompt success predictions accurately aggregate to reflect overall benchmark performance.

    \subsubsection{Model Selection}
    In \Cref{tab:sp_ms}, we present a comparison of our method to previous art over the model selection task. Our method outperforms all previous art in the OOS setting. For the in-sample settings, our results are comparable to the best-performing training-based EmbedLLM approach. Furthermore, as shown in \Cref{sec:appendix_c} (\Cref{tab:routing_static}), our dynamic selection approach outperforms a static ``Best Source Performer'' baseline, notably so in in-sample settings.

    \subsubsection{Scalability Evaluation}

    In \Cref{fig:emb_create_time}, we present the computational time comparison of our method and previous art. The computation was executed on \texttt{Intel(R) Xeon(R)} CPU and \texttt{NVIDIA L40S} GPU. Our approach presents a negligible increase in computation time for an increasing number of models, which aligns with our expected asymptotically linear computation time described in \Cref{sec:scalability_analysis}.
    
    

\section{Discussion}
\label{sec:concl}

\subsection{Conclusions}
This work has introduced a novel approach for representing LLMs as linear operators within the prompts' semantic task space. To do so, we consider models' application on queries as semantic-space translations from input to output. We then define the representations as a linear mapping from input to the task's performance metric, and compute them via matrix inversion. The resulting representations then present a semantic interpretation of LLMs' application compared to the task's goal, and are utilized for performance predictions and subsequent model selection. 

The suggested approach is training-free and scales to increasing number of models and benchmarks with negligible computational cost. Moreover, as the computation is based on matrix inversion, it can be extended to encompass additional models and benchmarks without recomputing the pre-existing ones. Hereby, requiring minimal computation to apply to newly published models and benchmarks, which is a crucial property in the rapidly evolving model-benchmark ecosystem. Furthermore, our method achieves state-of-the-art or comparable results over various performance prediction and model selection tasks, and outperforms previous OOS baselines.



Model repositories' continuous and rapid expansion underscores the critical need for scalable solutions. Such solutions must allow users to efficiently estimate LLM properties and select models suitable for their specific purposes and operational constraints. We have demonstrated the efficiency and scalability of our embedding creation process, benefits directly attributable to the simplicity of the underlying linear operations. While our current work focuses on dense performance matrices, the underlying linear algebraic framework is amenable to future extensions, potentially incorporating matrix-completion techniques to handle scenarios with sparser performance data. These characteristics support efficient retrieval and search over extensive collections of models, datasets, and tasks.


Conceptually, we have advanced the understanding of LLM performance through the ``success hyperplane normal'' lens. Our embedding effectively captures the correlation between models' responses to various prompts and their corresponding measured performance. This semantic representation allows models, inputs, and task success criteria to be considered within a shared semantic framework, offering clearer insights into model-task alignment. Furthermore, maintaining a consistent representation space for models enables the parallel computation of our approach across distributed models' repositories.


This work lays the foundation for a continuous deployment process composed of benchmarking, embedding, and storing the embeddings as informative metadata of the models in the repository. This allows retrieval of LLMs based on predefined properties, thus supporting large-scale, accessible LLMs for practical applications. By utilizing the diversity of pre-existing models and benchmarks, we enable the identification of suitable models while minimizing the effort and environmental impact associated with training new models.


\subsection{Future Work}
Our approach to embedding LLMs in a task-oriented space opens several promising avenues for future research. A possible direction is to examine properties other than success, such as safety, efficiency, stylistic alignment, and personalization. Models would then be retrieved based on aggregated criteria, which enables the consideration of multiple desired properties. Similarly, we can extend our approach to encompass multiple tasks and produce representations that better interpret models' applications by considering distinct success metrics and corresponding behaviors.
Finally, we propose applying our method to dynamic task routing in multi-agent systems, where its training-free, scalable, and interpretable nature is uniquely suited for embedding the output of one agent to select the next, enabling more adaptive and explainable problem-solving pipelines of specialized LLMs.


\section{Limitations}
\label{sec:limit}
This work proposes representing LLMs as embeddings in semantic task spaces, where the embedding encodes their corresponding performance. The models' embedding then corresponds to successfully answered queries and enables an efficient and explainable search of well-performing models. We can then consider the semantic vectors of models as representing their corresponding semantic translation between inputs and outputs. However, our suggested representation only regards a single task and its corresponding information, which may be insufficient to represent the complex semantic translation of LLMs. Moreover, our representation does not consider the models' architectures or inference complexity in any way, and future work could extend it to also consider model efficiency in addition to performance.

\clearpage

\bibliography{main}

\begin{thebibliography}{67}
\providecommand{\natexlab}[1]{#1}

\bibitem[{Alzahrani et~al.(2024)Alzahrani, Alyahya, Alnumay, AlRashed, Alsubaie, Almushayqih, Mirza, Alotaibi, Al-Twairesh, Alowisheq, Bari, and Khan}]{alzahrani-etal-2024-benchmarks}
Norah Alzahrani, Hisham Alyahya, Yazeed Alnumay, Sultan AlRashed, Shaykhah Alsubaie, Yousef Almushayqih, Faisal Mirza, Nouf Alotaibi, Nora Al-Twairesh, Areeb Alowisheq, M~Saiful Bari, and Haidar Khan. 2024.
\newblock \href {https://doi.org/10.18653/v1/2024.acl-long.744} {When benchmarks are targets: Revealing the sensitivity of large language model leaderboards}.
\newblock In \emph{Proceedings of the 62nd Annual Meeting of the Association for Computational Linguistics (Volume 1: Long Papers)}, pages 13787--13805, Bangkok, Thailand. Association for Computational Linguistics.

\bibitem[{Amini et~al.(2019)Amini, Gabriel, Lin, Koncel-Kedziorski, Choi, and Hajishirzi}]{amini-etal-2019-mathqa}
Aida Amini, Saadia Gabriel, Shanchuan Lin, Rik Koncel-Kedziorski, Yejin Choi, and Hannaneh Hajishirzi. 2019.
\newblock \href {https://doi.org/10.18653/v1/N19-1245} {{M}ath{QA}: Towards interpretable math word problem solving with operation-based formalisms}.
\newblock In \emph{Proceedings of the 2019 Conference of the North {A}merican Chapter of the Association for Computational Linguistics: Human Language Technologies, Volume 1 (Long and Short Papers)}, pages 2357--2367, Minneapolis, Minnesota. Association for Computational Linguistics.

\bibitem[{Bao et~al.(2019)Bao, Li, Huang, Zhang, Zheng, Zamir, and Guibas}]{8803726}
Yajie Bao, Yang Li, Shao-Lun Huang, Lin Zhang, Lizhong Zheng, Amir Zamir, and Leonidas Guibas. 2019.
\newblock \href {https://doi.org/10.1109/ICIP.2019.8803726} {An information-theoretic approach to transferability in task transfer learning}.
\newblock In \emph{2019 IEEE International Conference on Image Processing (ICIP)}, pages 2309--2313.

\bibitem[{Bassignana et~al.(2022)Bassignana, M{\"u}ller-Eberstein, Zhang, and Plank}]{bassignana-etal-2022-evidence}
Elisa Bassignana, Max M{\"u}ller-Eberstein, Mike Zhang, and Barbara Plank. 2022.
\newblock \href {https://doi.org/10.18653/v1/2022.emnlp-main.283} {Evidence {\ensuremath{>}} intuition: Transferability estimation for encoder selection}.
\newblock In \emph{Proceedings of the 2022 Conference on Empirical Methods in Natural Language Processing}, pages 4218--4227, Abu Dhabi, United Arab Emirates. Association for Computational Linguistics.

\bibitem[{Ben-Israel and Greville(2003)}]{ben2003generalized}
Adi Ben-Israel and Thomas~N.E. Greville. 2003.
\newblock \emph{Generalized Inverses: Theory and Applications}, 2nd edition.
\newblock Springer, New York.

\bibitem[{Bisk et~al.(2020)Bisk, Zellers, Le~bras, Gao, and Choi}]{piqa}
Yonatan Bisk, Rowan Zellers, Ronan Le~bras, Jianfeng Gao, and Yejin Choi. 2020.
\newblock \href {https://doi.org/10.1609/aaai.v34i05.6239} {Piqa: Reasoning about physical commonsense in natural language}.
\newblock \emph{Proceedings of the AAAI Conference on Artificial Intelligence}, 34(05):7432--7439.

\bibitem[{Bjerhammar(1951)}]{bjerhammar1951application}
Arne Bjerhammar. 1951.
\newblock Application of calculus of matrices to method of least squares: with special reference to geodetic calculations.
\newblock \emph{Transactions of the Royal Institute of Technology, Stockholm}, 49:1--86.

\bibitem[{Bunch and Nielsen(1978)}]{bunch_updating_1978}
James~R. Bunch and Christopher~P. Nielsen. 1978.
\newblock \href {https://doi.org/10.1007/BF01397471} {Updating the singular value decomposition}.
\newblock \emph{Numerische Mathematik}, 31(2):111--129.

\bibitem[{Cao et~al.(2024)Cao, Cai, Zhang, Zou, and Lam}]{cao2024on}
Bowen Cao, Deng Cai, Zhisong Zhang, Yuexian Zou, and Wai Lam. 2024.
\newblock \href {https://openreview.net/forum?id=Mi853QaJx6} {On the worst prompt performance of large language models}.
\newblock In \emph{The Thirty-eighth Annual Conference on Neural Information Processing Systems}.

\bibitem[{Chang et~al.(2023)Chang, Wang, Wang, Wu, Yang, Zhu, Chen, Yi, Wang, Wang, Ye, Zhang, Chang, Yu, Yang, and Xie}]{chang2023surveyevaluationlargelanguage}
Yupeng Chang, Xu~Wang, Jindong Wang, Yuan Wu, Linyi Yang, Kaijie Zhu, Hao Chen, Xiaoyuan Yi, Cunxiang Wang, Yidong Wang, Wei Ye, Yue Zhang, Yi~Chang, Philip~S. Yu, Qiang Yang, and Xing Xie. 2023.
\newblock \href {https://arxiv.org/abs/2307.03109} {A survey on evaluation of large language models}.
\newblock \emph{Preprint}, arXiv:2307.03109.

\bibitem[{Chao et~al.(2024)Chao, Debenedetti, Robey, Andriushchenko, Croce, Sehwag, Dobriban, Flammarion, Pappas, Tram{\`e}r, Hassani, and Wong}]{chao2024jailbreakbench}
Patrick Chao, Edoardo Debenedetti, Alexander Robey, Maksym Andriushchenko, Francesco Croce, Vikash Sehwag, Edgar Dobriban, Nicolas Flammarion, George~J. Pappas, Florian Tram{\`e}r, Hamed Hassani, and Eric Wong. 2024.
\newblock \href {https://openreview.net/forum?id=urjPCYZt0I} {Jailbreakbench: An open robustness benchmark for jailbreaking large language models}.
\newblock In \emph{The Thirty-eight Conference on Neural Information Processing Systems Datasets and Benchmarks Track}.

\bibitem[{Chen et~al.(2025)Chen, Huang, Gao, Wang, Yang, and Ji}]{chen2025scaling}
Yangyi Chen, Binxuan Huang, Yifan Gao, Zhengyang Wang, Jingfeng Yang, and Heng Ji. 2025.
\newblock \href {https://openreview.net/forum?id=BDisxnHzRL} {Scaling laws for predicting downstream performance in {LLM}s}.

\bibitem[{Chung et~al.(2024)Chung, Hou, Longpre, Zoph, Tay, Fedus, Li, Wang, Dehghani, Brahma, Webson, Gu, Dai, Suzgun, Chen, Chowdhery, Castro-Ros, Pellat, Robinson, Valter, Narang, Mishra, Yu, Zhao, Huang, Dai, Yu, Petrov, Chi, Dean, Devlin, Roberts, Zhou, Le, and Wei}]{JMLR:v25:23-0870}
Hyung~Won Chung, Le~Hou, Shayne Longpre, Barret Zoph, Yi~Tay, William Fedus, Yunxuan Li, Xuezhi Wang, Mostafa Dehghani, Siddhartha Brahma, Albert Webson, Shixiang~Shane Gu, Zhuyun Dai, Mirac Suzgun, Xinyun Chen, Aakanksha Chowdhery, Alex Castro-Ros, Marie Pellat, Kevin Robinson, and 16 others. 2024.
\newblock \href {http://jmlr.org/papers/v25/23-0870.html} {Scaling instruction-finetuned language models}.
\newblock \emph{Journal of Machine Learning Research}, 25(70):1--53.

\bibitem[{Cobbe et~al.(2021)Cobbe, Kosaraju, Bavarian, Chen, Jun, Kaiser, Plappert, Tworek, Hilton, Nakano, Hesse, and Schulman}]{cobbe2021trainingverifierssolvemath}
Karl Cobbe, Vineet Kosaraju, Mohammad Bavarian, Mark Chen, Heewoo Jun, Lukasz Kaiser, Matthias Plappert, Jerry Tworek, Jacob Hilton, Reiichiro Nakano, Christopher Hesse, and John Schulman. 2021.
\newblock \href {https://arxiv.org/abs/2110.14168} {Training verifiers to solve math word problems}.
\newblock \emph{Preprint}, arXiv:2110.14168.

\bibitem[{Devlin et~al.(2019)Devlin, Chang, Lee, and Toutanova}]{devlin-etal-2019-bert}
Jacob Devlin, Ming-Wei Chang, Kenton Lee, and Kristina Toutanova. 2019.
\newblock \href {https://doi.org/10.18653/v1/N19-1423} {{BERT}: Pre-training of deep bidirectional transformers for language understanding}.
\newblock In \emph{Proceedings of the 2019 Conference of the North {A}merican Chapter of the Association for Computational Linguistics: Human Language Technologies, Volume 1 (Long and Short Papers)}, pages 4171--4186, Minneapolis, Minnesota. Association for Computational Linguistics.

\bibitem[{Drori et~al.(2019)Drori, Liu, Nian, Koorathota, Li, Moretti, Freire, and Udell}]{drori2019automlusingmetadatalanguage}
Iddo Drori, Lu~Liu, Yi~Nian, Sharath~C. Koorathota, Jie~S. Li, Antonio~Khalil Moretti, Juliana Freire, and Madeleine Udell. 2019.
\newblock \href {https://arxiv.org/abs/1910.03698} {Automl using metadata language embeddings}.
\newblock \emph{Preprint}, arXiv:1910.03698.

\bibitem[{Eckart and Young(1936)}]{svdeckart1936approximation}
Carl Eckart and Gale Young. 1936.
\newblock The approximation of one matrix by another of lower rank.
\newblock \emph{Psychometrika}, 1(3):211--218.

\bibitem[{Etsenake and Nagappan(2024)}]{etsenake2024understandinghumanllmdynamicliterature}
Deborah Etsenake and Meiyappan Nagappan. 2024.
\newblock \href {https://arxiv.org/abs/2410.01026} {Understanding the human-llm dynamic: A literature survey of llm use in programming tasks}.
\newblock \emph{Preprint}, arXiv:2410.01026.

\bibitem[{Frieder et~al.(2024)Frieder, Berner, Petersen, and Lukasiewicz}]{frieder2024largelanguagemodelsmathematicians}
Simon Frieder, Julius Berner, Philipp Petersen, and Thomas Lukasiewicz. 2024.
\newblock \href {https://arxiv.org/abs/2312.04556} {Large language models for mathematicians}.
\newblock \emph{Preprint}, arXiv:2312.04556.

\bibitem[{Garg et~al.(2022)Garg, Balakrishnan, Lipton, Neyshabur, and Sedghi}]{garg2022leveraging}
Saurabh Garg, Sivaraman Balakrishnan, Zachary~Chase Lipton, Behnam Neyshabur, and Hanie Sedghi. 2022.
\newblock \href {https://openreview.net/forum?id=o_HsiMPYh_x} {Leveraging unlabeled data to predict out-of-distribution performance}.
\newblock In \emph{International Conference on Learning Representations}.

\bibitem[{Hager(1989)}]{hager1989updating}
William~W Hager. 1989.
\newblock Updating the inverse of a matrix.
\newblock \emph{SIAM review}, 31(2):221--239.

\bibitem[{Hendrycks et~al.(2021)Hendrycks, Burns, Basart, Zou, Mazeika, Song, and Steinhardt}]{hendrycks2021measuring}
Dan Hendrycks, Collin Burns, Steven Basart, Andy Zou, Mantas Mazeika, Dawn Song, and Jacob Steinhardt. 2021.
\newblock \href {https://openreview.net/forum?id=d7KBjmI3GmQ} {Measuring massive multitask language understanding}.
\newblock In \emph{International Conference on Learning Representations}.

\bibitem[{Hoerl and Kennard(1970{\natexlab{a}})}]{hoerl1970ridge2}
Arthur~E. Hoerl and Robert~W. Kennard. 1970{\natexlab{a}}.
\newblock Ridge regression: Applications to nonorthogonal problems.
\newblock \emph{Technometrics}, 12(1):69--82.

\bibitem[{Hoerl and Kennard(1970{\natexlab{b}})}]{hoerl1970ridge1}
Arthur~E. Hoerl and Robert~W. Kennard. 1970{\natexlab{b}}.
\newblock Ridge regression: Biased estimation for nonorthogonal problems.
\newblock \emph{Technometrics}, 12(1):55--67.

\bibitem[{Hu et~al.(2022)Hu, yelong shen, Wallis, Allen-Zhu, Li, Wang, Wang, and Chen}]{hu2022lora}
Edward~J Hu, yelong shen, Phillip Wallis, Zeyuan Allen-Zhu, Yuanzhi Li, Shean Wang, Lu~Wang, and Weizhu Chen. 2022.
\newblock \href {https://openreview.net/forum?id=nZeVKeeFYf9} {Lo{RA}: Low-rank adaptation of large language models}.
\newblock In \emph{International Conference on Learning Representations}.

\bibitem[{Huang et~al.(2024)Huang, Liu, Lin, Pang, Du, and Lin}]{huang2024lorahub}
Chengsong Huang, Qian Liu, Bill~Yuchen Lin, Tianyu Pang, Chao Du, and Min Lin. 2024.
\newblock \href {https://openreview.net/forum?id=TrloAXEJ2B} {Lorahub: Efficient cross-task generalization via dynamic lo{RA} composition}.
\newblock In \emph{First Conference on Language Modeling}.

\bibitem[{Jawahar et~al.(2024)Jawahar, Abdul-Mageed, Lakshmanan, and Ding}]{jawahar-etal-2024-llm}
Ganesh Jawahar, Muhammad Abdul-Mageed, Laks Lakshmanan, and Dujian Ding. 2024.
\newblock \href {https://doi.org/10.18653/v1/2024.findings-acl.627} {{LLM} performance predictors are good initializers for architecture search}.
\newblock In \emph{Findings of the Association for Computational Linguistics: ACL 2024}, pages 10540--10560, Bangkok, Thailand. Association for Computational Linguistics.

\bibitem[{Jiang et~al.(2024)Jiang, Wang, Shen, Kim, and Kim}]{jiang2024surveylargelanguagemodels}
Juyong Jiang, Fan Wang, Jiasi Shen, Sungju Kim, and Sunghun Kim. 2024.
\newblock \href {https://arxiv.org/abs/2406.00515} {A survey on large language models for code generation}.
\newblock \emph{Preprint}, arXiv:2406.00515.

\bibitem[{Jimenez et~al.(2024)Jimenez, Yang, Wettig, Yao, Pei, Press, and Narasimhan}]{jimenez2024swebenchlanguagemodelsresolve}
Carlos~E. Jimenez, John Yang, Alexander Wettig, Shunyu Yao, Kexin Pei, Ofir Press, and Karthik Narasimhan. 2024.
\newblock \href {https://arxiv.org/abs/2310.06770} {Swe-bench: Can language models resolve real-world github issues?}
\newblock \emph{Preprint}, arXiv:2310.06770.

\bibitem[{Jin et~al.(2024)Jin, Wan, Leaman, Tian, Wang, Yang, Wang, Xiong, Lai, Zhu, Hou, Sarfo-Gyamfi, Zhang, Gilson, Bhasuran, He, Zhang, Sun, Weng, Summers, Chen, Peng, and Lu}]{jin2024demystifyinglargelanguagemodels}
Qiao Jin, Nicholas Wan, Robert Leaman, Shubo Tian, Zhizheng Wang, Yifan Yang, Zifeng Wang, Guangzhi Xiong, Po-Ting Lai, Qingqing Zhu, Benjamin Hou, Maame Sarfo-Gyamfi, Gongbo Zhang, Aidan Gilson, Balu Bhasuran, Zhe He, Aidong Zhang, Jimeng Sun, Chunhua Weng, and 4 others. 2024.
\newblock \href {https://arxiv.org/abs/2410.18856} {Demystifying large language models for medicine: A primer}.
\newblock \emph{Preprint}, arXiv:2410.18856.

\bibitem[{Li et~al.(2024{\natexlab{a}})Li, Dong, Wang, Hu, Zuo, Lin, Qiao, and Shao}]{li-etal-2024-salad}
Lijun Li, Bowen Dong, Ruohui Wang, Xuhao Hu, Wangmeng Zuo, Dahua Lin, Yu~Qiao, and Jing Shao. 2024{\natexlab{a}}.
\newblock \href {https://doi.org/10.18653/v1/2024.findings-acl.235} {{SALAD}-bench: A hierarchical and comprehensive safety benchmark for large language models}.
\newblock In \emph{Findings of the Association for Computational Linguistics: ACL 2024}, pages 3923--3954, Bangkok, Thailand. Association for Computational Linguistics.

\bibitem[{Li et~al.(2024{\natexlab{b}})Li, Huang, Zhuang, Shi, Cai, Xu, Wang, Zhang, Ke, and Cai}]{li2024scilitllmadaptllmsscientific}
Sihang Li, Jin Huang, Jiaxi Zhuang, Yaorui Shi, Xiaochen Cai, Mingjun Xu, Xiang Wang, Linfeng Zhang, Guolin Ke, and Hengxing Cai. 2024{\natexlab{b}}.
\newblock \href {https://arxiv.org/abs/2408.15545} {Scilitllm: How to adapt llms for scientific literature understanding}.
\newblock \emph{Preprint}, arXiv:2408.15545.

\bibitem[{Lin et~al.(2021)Lin, Hilton, and Evans}]{Lin2021TruthfulQAMH}
Stephanie~C. Lin, Jacob Hilton, and Owain Evans. 2021.
\newblock \href {https://api.semanticscholar.org/CorpusID:237532606} {Truthfulqa: Measuring how models mimic human falsehoods}.
\newblock In \emph{Annual Meeting of the Association for Computational Linguistics}.

\bibitem[{Liu et~al.(2020)Liu, Cui, Liu, Huang, Wang, and Zhang}]{ijcai2020p501}
Jian Liu, Leyang Cui, Hanmeng Liu, Dandan Huang, Yile Wang, and Yue Zhang. 2020.
\newblock \href {https://doi.org/10.24963/ijcai.2020/501} {Logiqa: A challenge dataset for machine reading comprehension with logical reasoning}.
\newblock In \emph{Proceedings of the Twenty-Ninth International Joint Conference on Artificial Intelligence, {IJCAI-20}}, pages 3622--3628. International Joint Conferences on Artificial Intelligence Organization.
\newblock Main track.

\bibitem[{Liu et~al.(2024)Liu, Yu, Zhang, Xu, Lei, Lai, Gu, Ding, Men, Yang, Zhang, Deng, Zeng, Du, Zhang, Shen, Zhang, Su, Sun, Huang, Dong, and Tang}]{liu2024agentbench}
Xiao Liu, Hao Yu, Hanchen Zhang, Yifan Xu, Xuanyu Lei, Hanyu Lai, Yu~Gu, Hangliang Ding, Kaiwen Men, Kejuan Yang, Shudan Zhang, Xiang Deng, Aohan Zeng, Zhengxiao Du, Chenhui Zhang, Sheng Shen, Tianjun Zhang, Yu~Su, Huan Sun, and 3 others. 2024.
\newblock \href {https://openreview.net/forum?id=zAdUB0aCTQ} {Agentbench: Evaluating {LLM}s as agents}.
\newblock In \emph{The Twelfth International Conference on Learning Representations}.

\bibitem[{Miao et~al.(2020)Miao, Liang, and Su}]{miao-etal-2020-diverse}
Shen-yun Miao, Chao-Chun Liang, and Keh-Yih Su. 2020.
\newblock \href {https://doi.org/10.18653/v1/2020.acl-main.92} {A diverse corpus for evaluating and developing {E}nglish math word problem solvers}.
\newblock In \emph{Proceedings of the 58th Annual Meeting of the Association for Computational Linguistics}, pages 975--984, Online. Association for Computational Linguistics.

\bibitem[{Mikolov et~al.(2013)Mikolov, Chen, Corrado, and Dean}]{Mikolov2013EfficientEO}
Tomas Mikolov, Kai Chen, Gregory~S. Corrado, and Jeffrey Dean. 2013.
\newblock \href {https://api.semanticscholar.org/CorpusID:5959482} {Efficient estimation of word representations in vector space}.
\newblock In \emph{International Conference on Learning Representations}.

\bibitem[{Miller(2024)}]{miller2024addingerrorbarsevals}
Evan Miller. 2024.
\newblock \href {https://arxiv.org/abs/2411.00640} {Adding error bars to evals: A statistical approach to language model evaluations}.
\newblock \emph{Preprint}, arXiv:2411.00640.

\bibitem[{Moore(1920)}]{moore1920reciprocal}
Eliakim~Hastings Moore. 1920.
\newblock On the reciprocal of the general algebraic matrix.
\newblock \emph{Bulletin of the American Mathematical Society}, 26(9):394--395.

\bibitem[{Pacchiardi et~al.(2025)Pacchiardi, Cheke, and Hernandez-Orallo}]{pacchiardi2025}
Lorenzo Pacchiardi, Lucy~G Cheke, and Jose Hernandez-Orallo. 2025.
\newblock \href {https://openreview.net/forum?id=UoWslU6hsX} {100 instances is all you need: predicting {LLM} success by testing on a few instances}.

\bibitem[{Pal et~al.(2022)Pal, Umapathi, and Sankarasubbu}]{pmlr-v174-pal22a}
Ankit Pal, Logesh~Kumar Umapathi, and Malaikannan Sankarasubbu. 2022.
\newblock \href {https://proceedings.mlr.press/v174/pal22a.html} {Medmcqa: A large-scale multi-subject multi-choice dataset for medical domain question answering}.
\newblock In \emph{Proceedings of the Conference on Health, Inference, and Learning}, volume 174 of \emph{Proceedings of Machine Learning Research}, pages 248--260. PMLR.

\bibitem[{Papadopoulos et~al.(2007)Papadopoulos, Vovk, and Gammerman}]{4410411}
Harris Papadopoulos, Volodya Vovk, and Alex Gammerman. 2007.
\newblock \href {https://doi.org/10.1109/ICTAI.2007.47} {Conformal prediction with neural networks}.
\newblock In \emph{19th IEEE International Conference on Tools with Artificial Intelligence(ICTAI 2007)}, volume~2, pages 388--395.

\bibitem[{Parmar et~al.(2023)Parmar, Varshney, Patel, Mashetty, Luo, Mitra, and Baral}]{parmar2023logicbench}
Mihir Parmar, Neeraj Varshney, Nisarg Patel, Santosh Mashetty, Man Luo, Arindam Mitra, and Chitta Baral. 2023.
\newblock \href {https://openreview.net/forum?id=7NR2ZVzZxx} {Logicbench: A benchmark for evaluation of logical reasoning}.

\bibitem[{Penrose(1955)}]{penrose1955generalized}
Roger Penrose. 1955.
\newblock A generalized inverse for matrices.
\newblock \emph{Mathematical Proceedings of the Cambridge Philosophical Society}, 51(3):406--413.

\bibitem[{Perlitz et~al.(2024)Perlitz, Bandel, Gera, Arviv, Ein-Dor, Shnarch, Slonim, Shmueli-Scheuer, and Choshen}]{perlitz-etal-2024-efficient}
Yotam Perlitz, Elron Bandel, Ariel Gera, Ofir Arviv, Liat Ein-Dor, Eyal Shnarch, Noam Slonim, Michal Shmueli-Scheuer, and Leshem Choshen. 2024.
\newblock \href {https://doi.org/10.18653/v1/2024.naacl-long.139} {Efficient benchmarking (of language models)}.
\newblock In \emph{Proceedings of the 2024 Conference of the North American Chapter of the Association for Computational Linguistics: Human Language Technologies (Volume 1: Long Papers)}, pages 2519--2536, Mexico City, Mexico. Association for Computational Linguistics.

\bibitem[{Pezeshkpour and Hruschka(2024)}]{pezeshkpour-hruschka-2024-large}
Pouya Pezeshkpour and Estevam Hruschka. 2024.
\newblock \href {https://doi.org/10.18653/v1/2024.findings-naacl.130} {Large language models sensitivity to the order of options in multiple-choice questions}.
\newblock In \emph{Findings of the Association for Computational Linguistics: NAACL 2024}, pages 2006--2017, Mexico City, Mexico. Association for Computational Linguistics.

\bibitem[{Reimers and Gurevych(2019)}]{reimers-2019-sentence-bert}
Nils Reimers and Iryna Gurevych. 2019.
\newblock \href {http://arxiv.org/abs/1908.10084} {Sentence-bert: Sentence embeddings using siamese bert-networks}.
\newblock In \emph{Proceedings of the 2019 Conference on Empirical Methods in Natural Language Processing}. Association for Computational Linguistics.

\bibitem[{Rein et~al.(2024)Rein, Hou, Stickland, Petty, Pang, Dirani, Michael, and Bowman}]{rein2024gpqa}
David Rein, Betty~Li Hou, Asa~Cooper Stickland, Jackson Petty, Richard~Yuanzhe Pang, Julien Dirani, Julian Michael, and Samuel~R. Bowman. 2024.
\newblock \href {https://openreview.net/forum?id=Ti67584b98} {{GPQA}: A graduate-level google-proof q\&a benchmark}.
\newblock In \emph{First Conference on Language Modeling}.

\bibitem[{Sap et~al.(2019)Sap, Rashkin, Chen, Le~Bras, and Choi}]{sap-etal-2019-social}
Maarten Sap, Hannah Rashkin, Derek Chen, Ronan Le~Bras, and Yejin Choi. 2019.
\newblock \href {https://doi.org/10.18653/v1/D19-1454} {Social {IQ}a: Commonsense reasoning about social interactions}.
\newblock In \emph{Proceedings of the 2019 Conference on Empirical Methods in Natural Language Processing and the 9th International Joint Conference on Natural Language Processing (EMNLP-IJCNLP)}, pages 4463--4473, Hong Kong, China. Association for Computational Linguistics.

\bibitem[{Schroff et~al.(2015)Schroff, Kalenichenko, and Philbin}]{Schroff_2015_CVPR}
Florian Schroff, Dmitry Kalenichenko, and James Philbin. 2015.
\newblock Facenet: A unified embedding for face recognition and clustering.
\newblock In \emph{Proceedings of the IEEE Conference on Computer Vision and Pattern Recognition (CVPR)}.

\bibitem[{Schulz(1933)}]{Schulz1933}
G.~Schulz. 1933.
\newblock \href {https://doi.org/10.1002/zamm.19330130109} {Iterative {B}erechnung der reziproken {M}atrix}.
\newblock \emph{Zeitschrift f{\"u}r Angewandte Mathematik und Mechanik (ZAMM)}, 13(1):57--59.

\bibitem[{Song et~al.(2020)Song, Tan, Qin, Lu, and Liu}]{NEURIPS2020_c3a690be}
Kaitao Song, Xu~Tan, Tao Qin, Jianfeng Lu, and Tie-Yan Liu. 2020.
\newblock \href {https://proceedings.neurips.cc/paper_files/paper/2020/file/c3a690be93aa602ee2dc0ccab5b7b67e-Paper.pdf} {Mpnet: Masked and permuted pre-training for language understanding}.
\newblock In \emph{Advances in Neural Information Processing Systems}, volume~33, pages 16857--16867. Curran Associates, Inc.

\bibitem[{Suzgun et~al.(2023)Suzgun, Scales, Sch{\"a}rli, Gehrmann, Tay, Chung, Chowdhery, Le, Chi, Zhou, and Wei}]{suzgun-etal-2023-challenging}
Mirac Suzgun, Nathan Scales, Nathanael Sch{\"a}rli, Sebastian Gehrmann, Yi~Tay, Hyung~Won Chung, Aakanksha Chowdhery, Quoc Le, Ed~Chi, Denny Zhou, and Jason Wei. 2023.
\newblock \href {https://doi.org/10.18653/v1/2023.findings-acl.824} {Challenging {BIG}-bench tasks and whether chain-of-thought can solve them}.
\newblock In \emph{Findings of the Association for Computational Linguistics: ACL 2023}, pages 13003--13051, Toronto, Canada. Association for Computational Linguistics.

\bibitem[{Talmor et~al.(2019)Talmor, Herzig, Lourie, and Berant}]{talmor-etal-2019-commonsenseqa}
Alon Talmor, Jonathan Herzig, Nicholas Lourie, and Jonathan Berant. 2019.
\newblock \href {https://doi.org/10.18653/v1/N19-1421} {{C}ommonsense{QA}: A question answering challenge targeting commonsense knowledge}.
\newblock In \emph{Proceedings of the 2019 Conference of the North {A}merican Chapter of the Association for Computational Linguistics: Human Language Technologies, Volume 1 (Long and Short Papers)}, pages 4149--4158, Minneapolis, Minnesota. Association for Computational Linguistics.

\bibitem[{Tikhonov(1943)}]{tikhonov1943stability}
Andrey~N. Tikhonov. 1943.
\newblock On the stability of inverse problems.
\newblock In \emph{Doklady Akademii Nauk SSSR}, volume~39, pages 195--198.

\bibitem[{Veli{\v{c}}kovi{\'c} et~al.(2022)Veli{\v{c}}kovi{\'c}, Badia, Budden, Pascanu, Banino, Dashevskiy, Hadsell, and Blundell}]{pmlr-v162-velickovic22a}
Petar Veli{\v{c}}kovi{\'c}, Adri{\`a}~Puigdom{\`e}nech Badia, David Budden, Razvan Pascanu, Andrea Banino, Misha Dashevskiy, Raia Hadsell, and Charles Blundell. 2022.
\newblock \href {https://proceedings.mlr.press/v162/velickovic22a.html} {The {CLRS} algorithmic reasoning benchmark}.
\newblock In \emph{Proceedings of the 39th International Conference on Machine Learning}, volume 162 of \emph{Proceedings of Machine Learning Research}, pages 22084--22102. PMLR.

\bibitem[{Wang et~al.(2021)Wang, Bao, Huang, Dong, and Wei}]{wang-etal-2021-minilmv2}
Wenhui Wang, Hangbo Bao, Shaohan Huang, Li~Dong, and Furu Wei. 2021.
\newblock \href {https://doi.org/10.18653/v1/2021.findings-acl.188} {{M}ini{LM}v2: Multi-head self-attention relation distillation for compressing pretrained transformers}.
\newblock In \emph{Findings of the Association for Computational Linguistics: ACL-IJCNLP 2021}, pages 2140--2151, Online. Association for Computational Linguistics.

\bibitem[{Wolf et~al.(2019)Wolf, Debut, Sanh, Chaumond, Delangue, Moi, Cistac, Rault, Louf, Funtowicz et~al.}]{wolf2019huggingface}
Thomas Wolf, Lysandre Debut, Victor Sanh, Julien Chaumond, Clement Delangue, Anthony Moi, Pierric Cistac, Tim Rault, R{\'e}mi Louf, Morgan Funtowicz, and 1 others. 2019.
\newblock Huggingface's transformers: State-of-the-art natural language processing.
\newblock \emph{arXiv preprint arXiv:1910.03771}.

\bibitem[{You et~al.(2021)You, Liu, Wang, and Long}]{pmlr-v139-you21b}
Kaichao You, Yong Liu, Jianmin Wang, and Mingsheng Long. 2021.
\newblock \href {https://proceedings.mlr.press/v139/you21b.html} {Logme: Practical assessment of pre-trained models for transfer learning}.
\newblock In \emph{Proceedings of the 38th International Conference on Machine Learning}, volume 139 of \emph{Proceedings of Machine Learning Research}, pages 12133--12143. PMLR.

\bibitem[{Yu et~al.(2024)Yu, Wang, Tu, Cao, Zhang-Li, Lv, Peng, Yao, Zhang, Li, Li, Zhang, Bai, Liu, Xin, Yun, GONG, Lin, Chen, Wu, Qi, Li, Guan, Zeng, Qi, Jin, Liu, Gu, Yao, Ding, Hou, Liu, Bin, Tang, and Li}]{yu2024kola}
Jifan Yu, Xiaozhi Wang, Shangqing Tu, Shulin Cao, Daniel Zhang-Li, Xin Lv, Hao Peng, Zijun Yao, Xiaohan Zhang, Hanming Li, Chunyang Li, Zheyuan Zhang, Yushi Bai, Yantao Liu, Amy Xin, Kaifeng Yun, Linlu GONG, Nianyi Lin, Jianhui Chen, and 16 others. 2024.
\newblock \href {https://openreview.net/forum?id=AqN23oqraW} {Ko{LA}: Carefully benchmarking world knowledge of large language models}.
\newblock In \emph{The Twelfth International Conference on Learning Representations}.

\bibitem[{Zhang et~al.(2024{\natexlab{a}})Zhang, Ding, Lyv, Wang, Yin, Zhang, Yu, Wang, Li, Xiang, Feng, Zhuang, Wang, Qin, Zhang, Zhang, Cui, Huang, Yan, Xu, Chen, Li, Fan, Xing, and Chen}]{zhang2024scientificlargelanguagemodels}
Qiang Zhang, Keyang Ding, Tianwen Lyv, Xinda Wang, Qingyu Yin, Yiwen Zhang, Jing Yu, Yuhao Wang, Xiaotong Li, Zhuoyi Xiang, Kehua Feng, Xiang Zhuang, Zeyuan Wang, Ming Qin, Mengyao Zhang, Jinlu Zhang, Jiyu Cui, Tao Huang, Pengju Yan, and 6 others. 2024{\natexlab{a}}.
\newblock \href {https://arxiv.org/abs/2401.14656} {Scientific large language models: A survey on biological \& chemical domains}.
\newblock \emph{Preprint}, arXiv:2401.14656.

\bibitem[{Zhang et~al.(2024{\natexlab{b}})Zhang, Lyu, Liu, and Ma}]{zhang-etal-2024-collaborative}
Qiyuan Zhang, Fuyuan Lyu, Xue Liu, and Chen Ma. 2024{\natexlab{b}}.
\newblock \href {https://doi.org/10.18653/v1/2024.emnlp-main.150} {Collaborative performance prediction for large language models}.
\newblock In \emph{Proceedings of the 2024 Conference on Empirical Methods in Natural Language Processing}, pages 2576--2596, Miami, Florida, USA. Association for Computational Linguistics.

\bibitem[{Zhang et~al.(2023)Zhang, Gong, Wu, Liu, and Zhou}]{zhang2023automlgptautomaticmachinelearning}
Shujian Zhang, Chengyue Gong, Lemeng Wu, Xingchao Liu, and Mingyuan Zhou. 2023.
\newblock \href {https://arxiv.org/abs/2305.02499} {Automl-gpt: Automatic machine learning with gpt}.
\newblock \emph{Preprint}, arXiv:2305.02499.

\bibitem[{Zhang et~al.(2024{\natexlab{c}})Zhang, Lei, Wu, Sun, Huang, Long, Liu, Lei, Tang, and Huang}]{zhang-etal-2024-safetybench}
Zhexin Zhang, Leqi Lei, Lindong Wu, Rui Sun, Yongkang Huang, Chong Long, Xiao Liu, Xuanyu Lei, Jie Tang, and Minlie Huang. 2024{\natexlab{c}}.
\newblock \href {https://doi.org/10.18653/v1/2024.acl-long.830} {{S}afety{B}ench: Evaluating the safety of large language models}.
\newblock In \emph{Proceedings of the 62nd Annual Meeting of the Association for Computational Linguistics (Volume 1: Long Papers)}, pages 15537--15553, Bangkok, Thailand. Association for Computational Linguistics.

\bibitem[{Zheng et~al.(2024)Zheng, Gan, Chen, Qi, Liang, and Yu}]{zheng2024largelanguagemodelsmedicine}
Yanxin Zheng, Wensheng Gan, Zefeng Chen, Zhenlian Qi, Qian Liang, and Philip~S. Yu. 2024.
\newblock \href {https://arxiv.org/abs/2405.13055} {Large language models for medicine: A survey}.
\newblock \emph{Preprint}, arXiv:2405.13055.

\bibitem[{Zhuang et~al.(2025)Zhuang, Wu, Wen, Li, Jiao, and Ramchandran}]{zhuang2025embedllm}
Richard Zhuang, Tianhao Wu, Zhaojin Wen, Andrew Li, Jiantao Jiao, and Kannan Ramchandran. 2025.
\newblock \href {https://openreview.net/forum?id=Fs9EabmQrJ} {Embed{LLM}: Learning compact representations of large language models}.
\newblock In \emph{The Thirteenth International Conference on Learning Representations}.

\bibitem[{Zhuo et~al.(2024)Zhuo, Zhang, Fang, Duan, Lin, and Chen}]{zhuo-etal-2024-prosa}
Jingming Zhuo, Songyang Zhang, Xinyu Fang, Haodong Duan, Dahua Lin, and Kai Chen. 2024.
\newblock \href {https://doi.org/10.18653/v1/2024.findings-emnlp.108} {{P}ro{SA}: Assessing and understanding the prompt sensitivity of {LLM}s}.
\newblock In \emph{Findings of the Association for Computational Linguistics: EMNLP 2024}, pages 1950--1976, Miami, Florida, USA. Association for Computational Linguistics.

\end{thebibliography}

\clearpage
\newpage
\appendix
\onecolumn

 \renewcommand{\thesection}{\Alph{section}}
 \section{Finding the optimal epsilon}
\label{sec:appendix_a}

We present the hyper-parameter tuning of $\varepsilon$ for each discussed setting.
Increasing $\varepsilon$ results in filtering directions, that correspond to lower variance of the prompt embeddings.
We can notice increasing epsilon can improve all metrics in the OOS scenario, until at some point it starts to decrease. This can be attributed to the cleaning of noise and the maintenance of dominant singular directions.
Another noticeable trend in two of the datasets is that the task of model selection, where false positives are more problematic than false negatives, requires a larger $\varepsilon$, than the one required for success prediction.

\begin{figure*}[h!]
 \centering
    \resizebox{0.88\linewidth}{!}{
        \begin{tabular}{cc}  
            \includegraphics[]{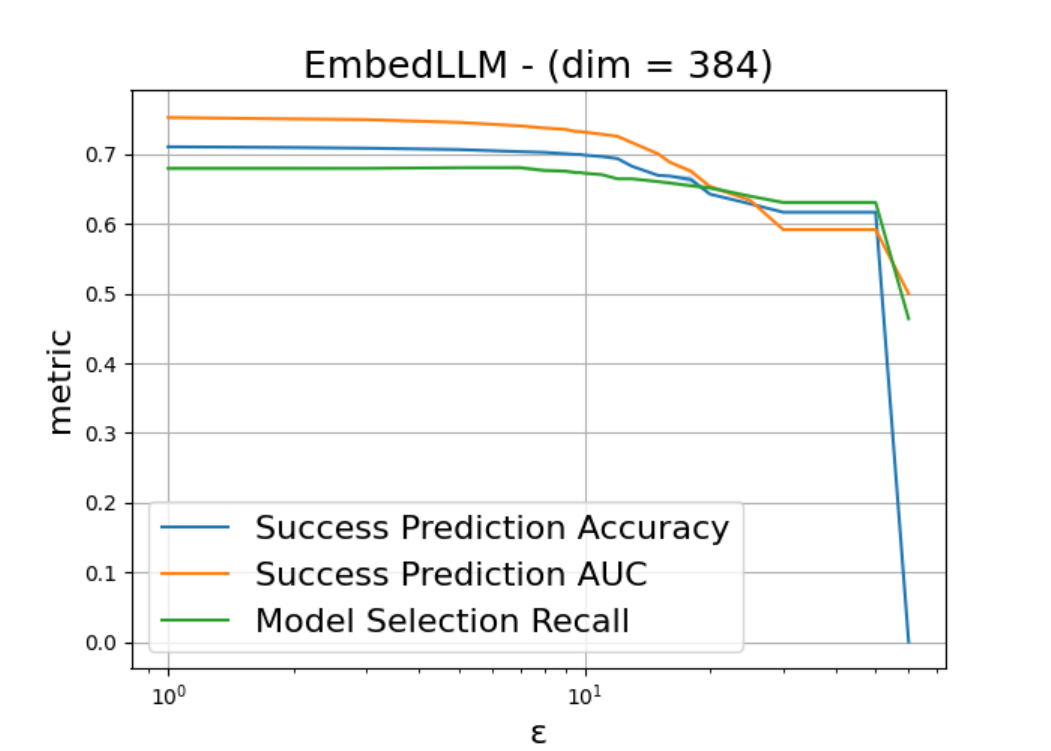}&
            \includegraphics[]{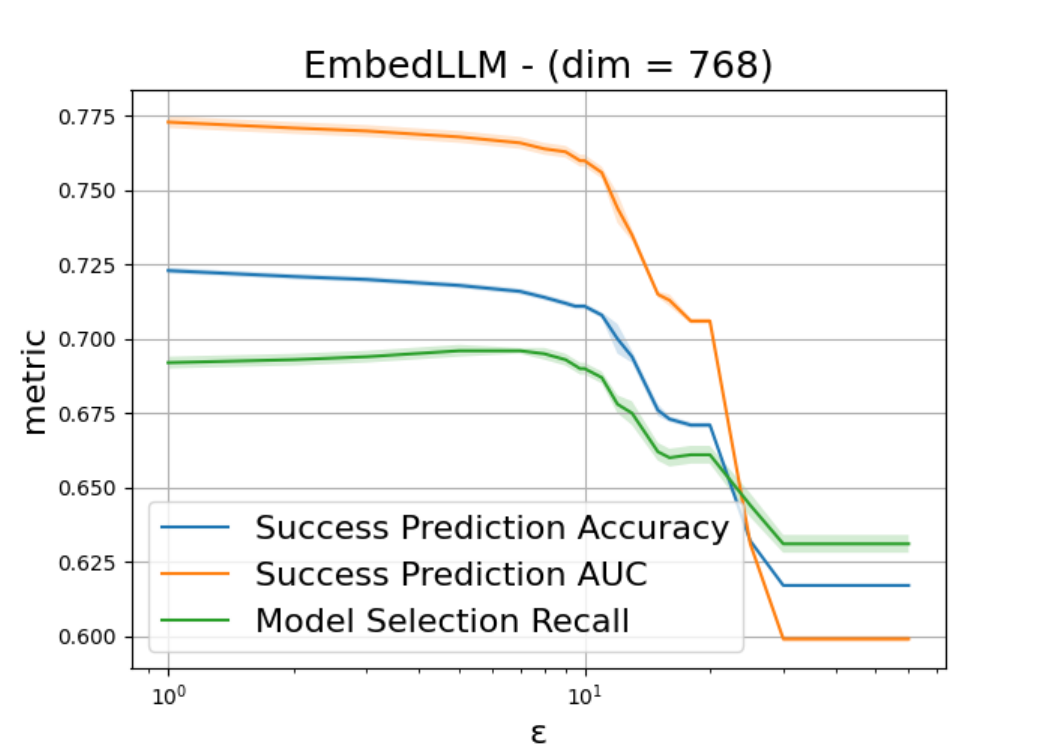}\\
            \includegraphics[]{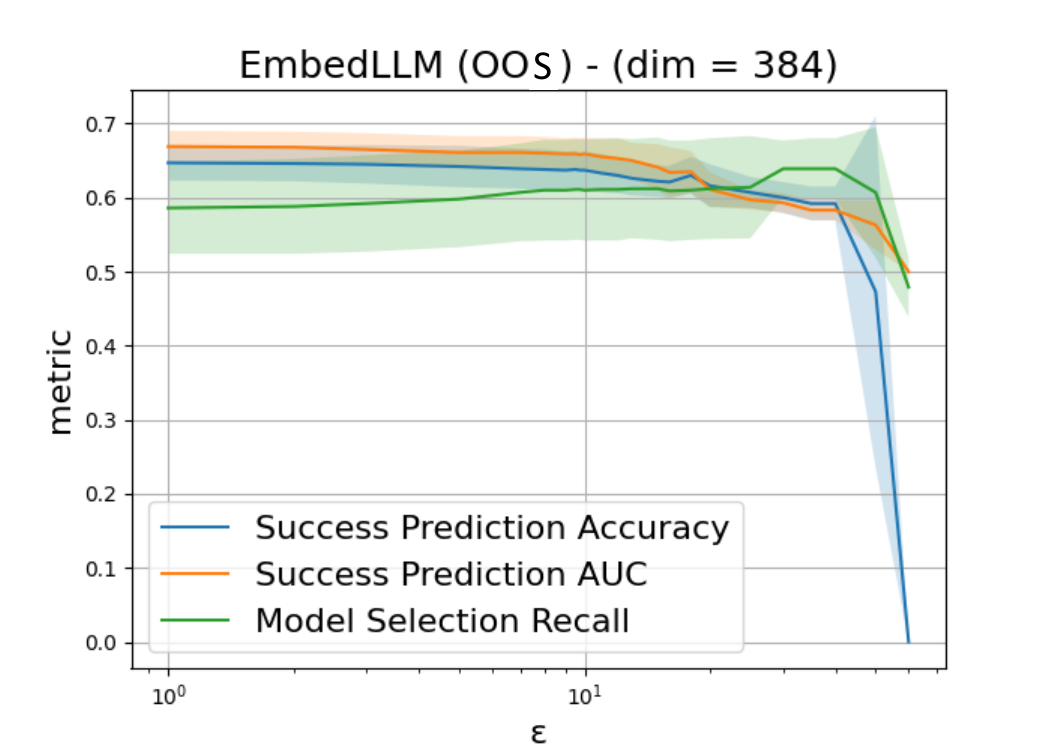}&
            \includegraphics[]{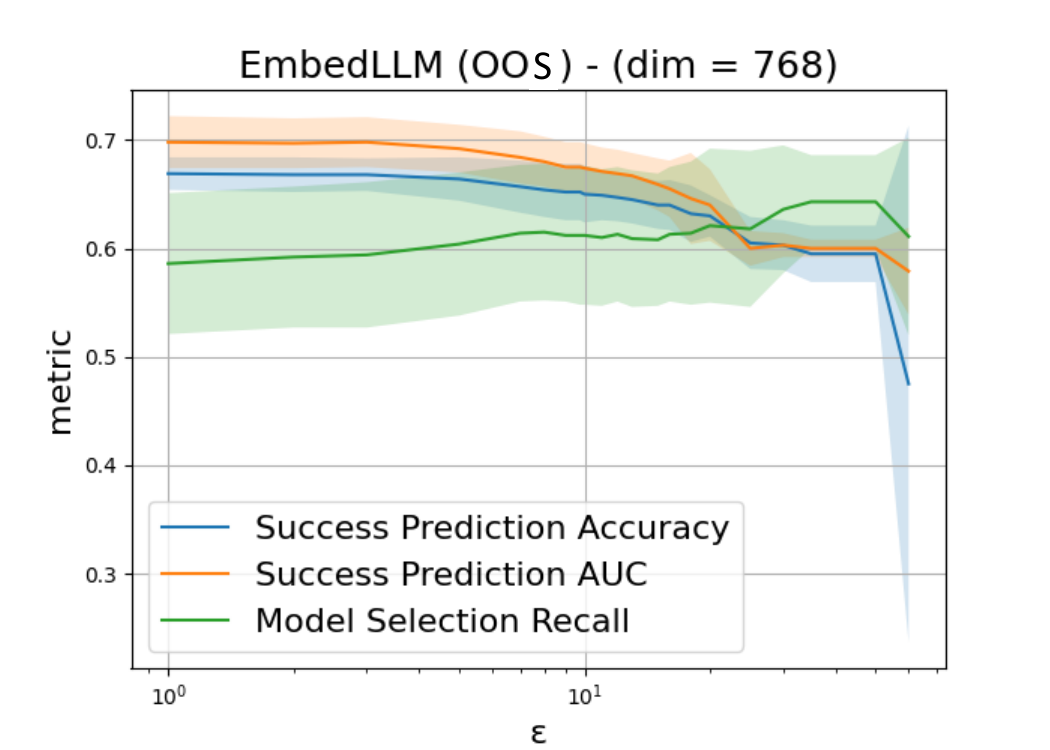}\\
            \includegraphics[]{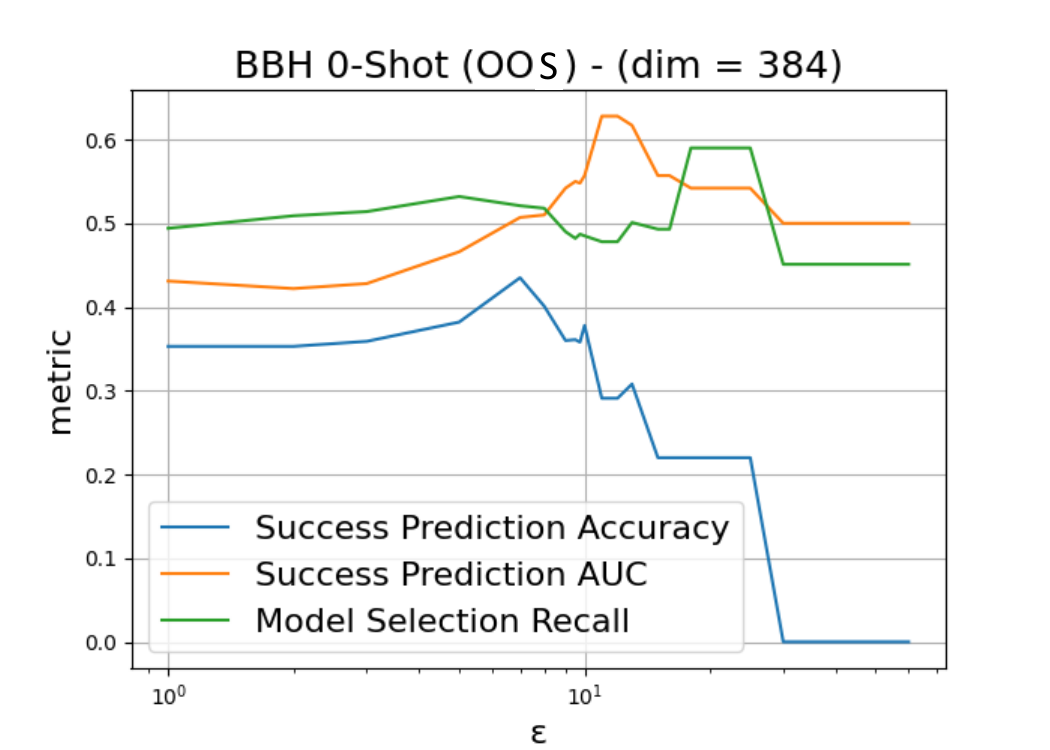}&
            \includegraphics[]{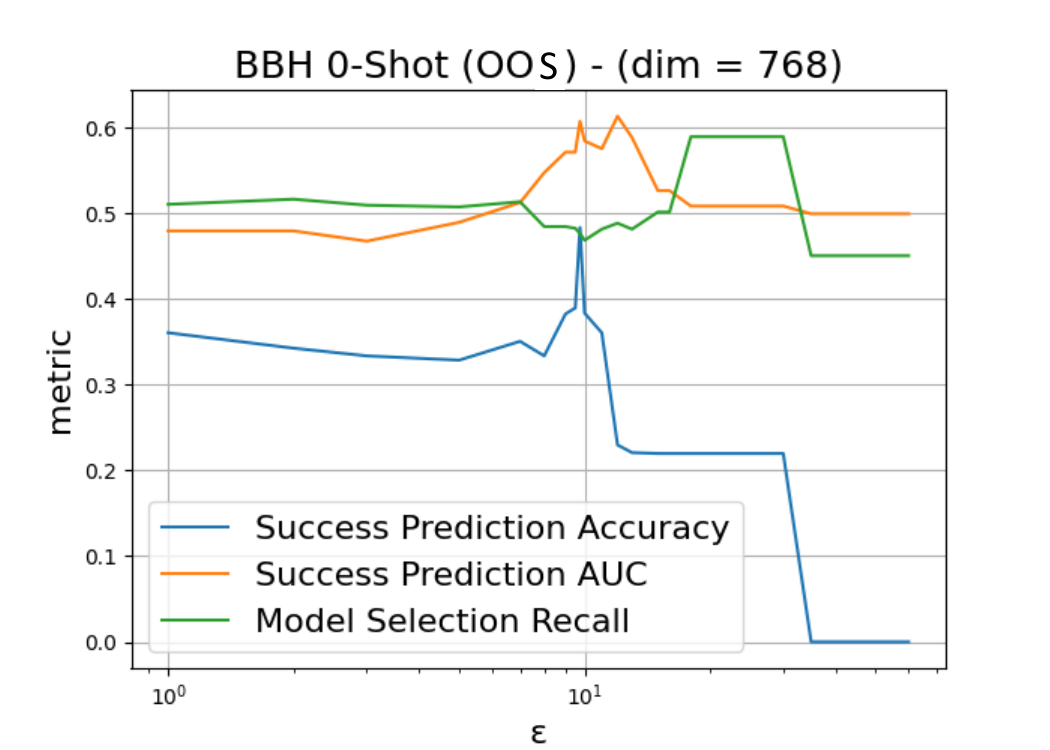}\\
            \includegraphics[]{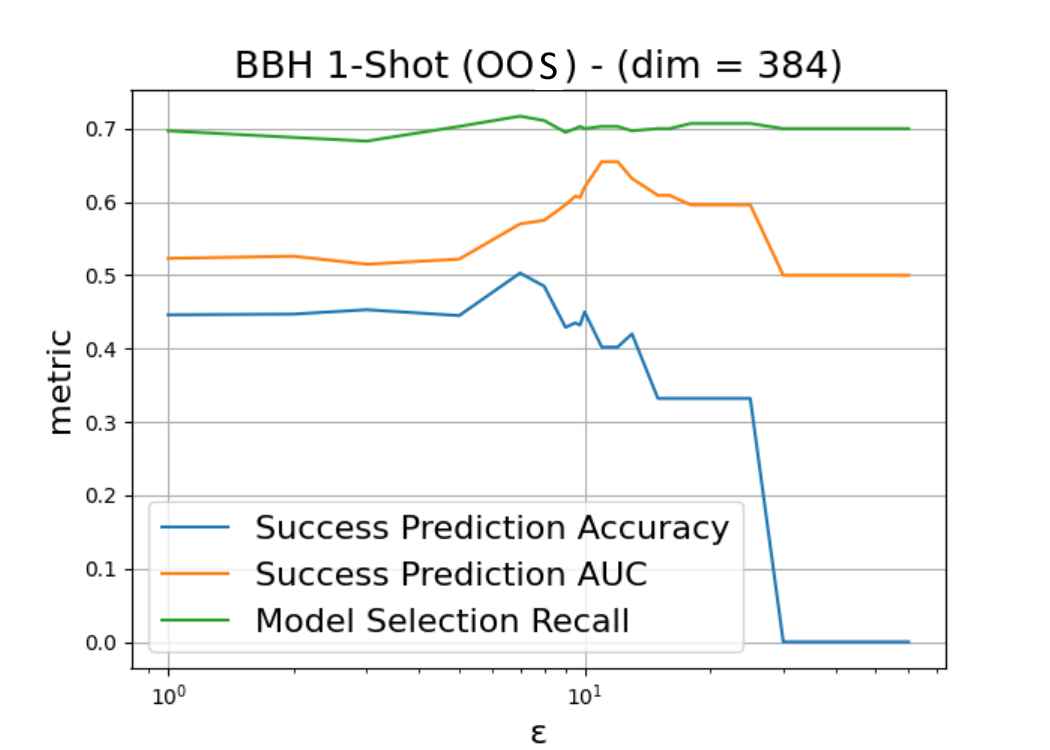}&
            \includegraphics[]{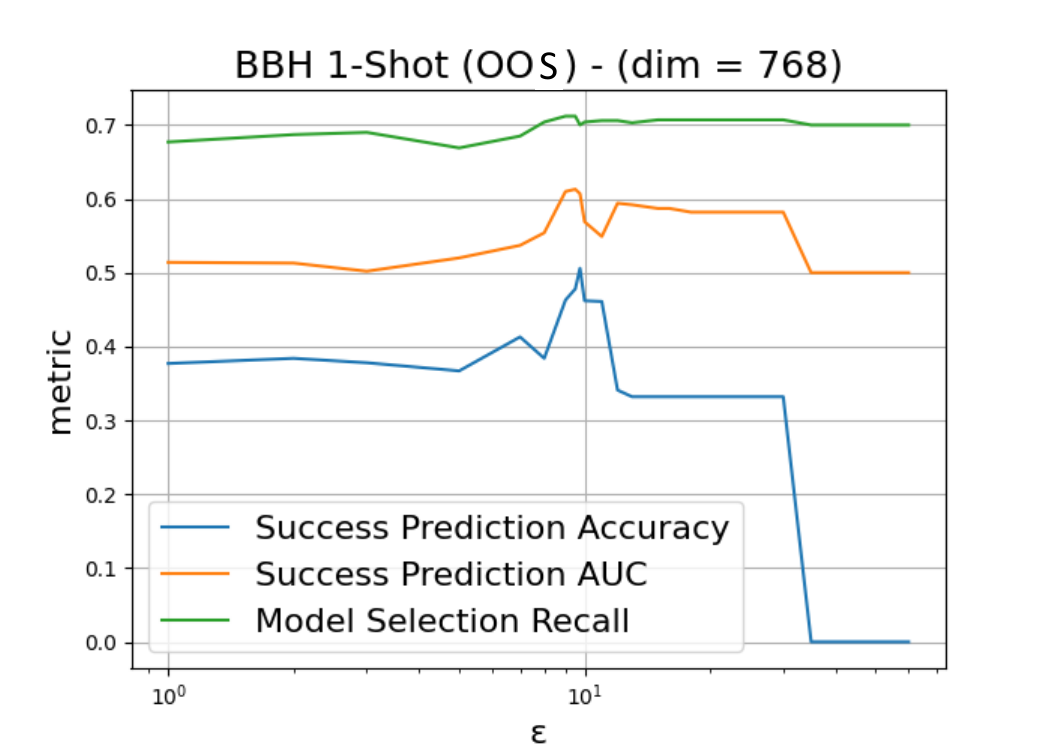}\\
        \end{tabular}
    }
    \caption{
    The effect of $\varepsilon$ on the performance metrics discussed in our work.
    }
    \label{fig:eps_ablation}
\end{figure*}

\clearpage
\section{Comparison with static selection}
\label{sec:appendix_c}
This section provides a comparative analysis of our dynamic model selection method against a static baseline termed the ``\textit{Best Source Performer}'' (BSP). The BSP baseline identifies the single LLM from the available pool that achieved the highest overall accuracy across all prompts within the defined source dataset ($\mathcal{D}_\text{src}$) for each specific evaluation environment. This pre-selected model is then used uniformly for all test prompts in that environment. This comparison serves to benchmark our per-prompt selection strategy against a strong, globally-optimized static choice based on performance over the known source data.

\Cref{tab:routing_static} presents the Accuracy and Recall metrics. The results generally show that our dynamic selection approach offers advantages, particularly in in-sample scenarios, while maintaining robust and competitive performance against the BSP baseline in OOS settings. This highlights the value of adaptive, per-prompt model selection.
\begin{table}[h]
\centering
\small
\caption{Comparison of our model selection method ('Ours') against a static 'Best Source Performer' baseline. The 'Best Source Performer' is the single model achieving the highest overall accuracy on the entire source prompt library (i.e., $\mathcal{D}_\text{src}$ from Section~\ref{sec:data_and_system}) for each environment. Performance is reported across different evaluation environments and prompt embedding dimensions for our method.}
\label{tab:routing_static} 
\setlength{\tabcolsep}{3pt}
\begin{tabular}{@{}llccc@{}}
\toprule\toprule
\multirow{2}{*}{Environment} & \multirow{2}{*}{Method} & \multirow{2}{*}{dim} & \multirow{2}{*}{Accuracy} & \multirow{2}{*}{Recall} \\
& & & & \\
\midrule

\multirow{3}{*}{\textbf{EmbedLLM}} 
& Best Source Performer & -- & $0.5759 \pm 0.0020$ & $0.6309 \pm 0.0026$ \\
& Ours                  & 384 & $\underline{0.6221 \pm 0.0020}$ & $\underline{0.6814 \pm 0.0027}$ \\
& Ours                  & 768 & $\mathbf{0.6355 \pm 0.0012}$ & $\mathbf{0.6961 \pm 0.0014}$ \\
\midrule

\multirow{3}{*}{\textbf{EmbedLLM (OOS)}} 
& Best Source Performer & -- & $\underline{0.5885 \pm 0.0444}$ & $\underline{0.6398 \pm 0.0414}$ \\
& Ours                  & 384 & $0.5879 \pm 0.0441$ & $0.6393 \pm 0.0413$ \\
& Ours                  & 768 & $\mathbf{0.5916 \pm 0.0436}$ & $\mathbf{0.6435 \pm 0.0435}$ \\
\midrule

\multirow{3}{*}{\textbf{BBH 0-shot (OOS)}} 
& Best Source Performer & -- & $\mathbf{0.2491}$ & $\mathbf{0.5896}$ \\
& Ours                  & 384 & $\mathbf{0.2491}$ & $\mathbf{0.5896}$ \\
& Ours                  & 768 & $\mathbf{0.2491}$ & $\mathbf{0.5896}$ \\
\midrule

\multirow{3}{*}{\textbf{BBH 1-shot (OOS)}} 
& Best Source Performer & -- & $0.3357$ & $0.7070$ \\
& Ours                  & 384 & $\mathbf{0.3405}$ & $\mathbf{0.7171}$ \\
& Ours                  & 768 & $\underline{0.3381}$ & $\underline{0.7119}$ \\
\bottomrule\bottomrule
\end{tabular}
\end{table}

\clearpage
\section{Algorithms for Incremental Updates}
\label{sec:appendix_incremental}
\subsection*{Adding New Models to the Repository}

Adding a new model is highly efficient as it does not require recomputing the expensive pseudoinverse of the prompt embedding matrix.
Calculating a new model's embedding merely requires \textit{a single matrix-multiplication operation.} The existing pseudoinverse, $(D_\text{src}^{+})^{\intercal}$, which encapsulates the structure of the source prompt space, is simply \textit{reused}, only this time with the measured performance of the new model on the source prompts.

\begin{algorithm}[H]
\caption{Incremental Addition of a New Model}
\begin{algorithmic}[1]
\State \textbf{Input:}
\State The precomputed pseudoinverse of the source prompt matrix, $(D_\text{src}^{+})^{\intercal} \in \mathbb{R}^{N \times d_\text{prompt}}$.
\State The performance vector $P_\text{new} \in \mathbb{R}^{1 \times N}$ for the new model $\mathcal{M}_\text{new}$ on the $N$ source prompts.
\Statex
\State \textbf{Procedure:}
\State Compute the embedding for the new model, $E(\mathcal{M})_{new}$, via a single matrix-vector multiplication:
\begin{equation*}
    E(\mathcal{M})_\text{new} = P_\text{new} \cdot (D_\text{src}^{+})^{\intercal}
\end{equation*}
\State Append the resulting vector $E(\mathcal{M})_\text{new} \in \mathbb{R}^{1 \times d_\text{prompt}}$ as a new row to the existing model embedding matrix $\mathbf{E(M)}$.
\end{algorithmic}
\end{algorithm}

\noindent \textbf{Computational Complexity:} The cost of this operation is dominated by the matrix-vector multiplication, which is $O(N \cdot d_\text{prompt})$. Since $d_\text{prompt}$ is fixed (e.g., 384 or 768), the complexity is linear in the number of source prompts, $N$. This cost is negligible compared to retraining-based approaches.

\subsection*{Adding New Source Prompts to the Library}

Adding new source prompts is more complex than adding models because it alters the source prompt matrix $\mathbf{D_{src}}$, invalidating the precomputed pseudoinverse $(\mathbf{D_{src}}^{+})^{\intercal}$. The most direct approach is to recompute the SVD of the new, larger prompt matrix, an operation with a complexity of $O(N \cdot d_{\text{prompt}}^2)$. For the dataset sizes explored in our scalability experiments (\Cref{fig:emb_create_time}), this direct recomputation was already so efficient that it resulted in nearly constant update times.

However, for large-scale systems where $N$ is exceptionally large, an even greater asymptotic efficiency can be achieved by using incremental update methods. One such approach is to compute the pseudoinverse via the normal equations. This involves inverting the square matrix $\mathbf{A} = \mathbf{D_{new}}^\intercal \mathbf{D_{new}}$. The Tikhonov regularization already present in our main method (\Cref{eq:sigma_prime_computation}) is equivalent to inverting $\mathbf{A} = \mathbf{D_{new}}^\intercal \mathbf{D_{new}} + 2\lambda\mathbf{I}$, which conveniently ensures the matrix is always invertible and well-conditioned. For this task, a classic iterative solver like the Newton-Schulz iteration~\cite{Schulz1933} can be used.

\begin{algorithm}[H]
\caption{Incremental Update of Model Embeddings via Newton-Schulz}
\label{alg:full_incremental_update}
\begin{algorithmic}[1]
\State \textbf{Input:}
\State Old source matrices: $\mathbf{D_{src}} \in \mathbb{R}^{N \times d_{\text{prompt}}}$, $\mathbf{P_{src}} \in \mathbb{R}^{M \times N}$.
\State Old computed inverse: $\mathbf{A_{src}}^{-1} = (\mathbf{D_{src}}^\intercal \mathbf{D_{src}} + 2\lambda\mathbf{I})^{-1} \in \mathbb{R}^{d_{\text{prompt}} \times d_{\text{prompt}}}$.
\State New data to add: $\mathbf{D_{added}} \in \mathbb{R}^{N_{add} \times d_{\text{prompt}}}$, $\mathbf{P_{added}} \in \mathbb{R}^{M \times N_{add}}$.
\State Iteration count for refinement, $k$.
\Statex
\State \textbf{Procedure:}
\State Concatenate matrices to form the new set:
\State $\mathbf{D_{new}} \gets \begin{bmatrix} \mathbf{D_{src}} \\ \mathbf{D_{added}} \end{bmatrix} \in \mathbb{R}^{(N+N_{add}) \times d_{\text{prompt}}}$
\State $\mathbf{P_{new}} \gets \begin{bmatrix} \mathbf{P_{src}} & \mathbf{P_{added}} \end{bmatrix} \in \mathbb{R}^{M \times (N+N_{add})}$
\Statex
\State Form the new matrix to be inverted:
\State $\mathbf{A_{new}} \gets \mathbf{D_{new}}^\intercal \mathbf{D_{new}} + 2\lambda\mathbf{I}$
\Statex
\State Use the previous inverse as a strong initial guess for the new inverse: $\mathbf{X}_0 \gets \mathbf{A_{src}}^{-1}$.
\For{$i = 0$ to $k-1$} 
    \State $\mathbf{X}_{i+1} \gets \mathbf{X}_i (2\mathbf{I} - \mathbf{A_{new}}\mathbf{X}_i)$ \Comment{Refine the inverse using Newton-Schulz iteration}
\EndFor
\State Let the converged inverse be $\mathbf{A_{new}}^{-1} \gets \mathbf{X}_k$.
\Statex
\State Compute the new pseudoinverse:
\State $\mathbf{D_{new}}^{+} \gets \mathbf{A_{new}}^{-1} \mathbf{D_{new}}^\intercal$
\Statex
\State Compute the final updated model embeddings:
\State $\mathbf{E(M)_{new}} \gets \mathbf{P_{new}} (\mathbf{D_{new}}^{+})^\intercal$
\Statex
\State \textbf{Output:} The updated model embedding matrix, $\mathbf{E(M)_{new}} \in \mathbb{R}^{M \times d_{\text{prompt}}}$.
\end{algorithmic}
\end{algorithm}

\noindent \textbf{Computational Complexity:} The dominant cost in Algorithm~\ref{alg:full_incremental_update} comes from the Newton-Schulz loop. Each iteration involves matrix multiplications of size ($d_{\text{prompt}} \times d_{\text{prompt}}$), leading to a complexity of $O(k \cdot d_{\text{prompt}}^3)$ for the loop. This is asymptotically more efficient than the full SVD recomputation ($O(N_{\text{new}} \cdot d_{\text{prompt}}^2)$) when the number of prompts $N_{\text{new}}$ is significantly larger than the embedding dimension $d_{\text{prompt}}$.

\noindent \textbf{Alternative Methods:} In addition to Newton-Schulz, there are other established methods for updating the pseudoinverse that are valid for our use, such as updating the SVD directly via rank-one updates~\cite{bunch_updating_1978} or using the Sherman-Morrison-Woodbury formula for low-rank updates~\cite{hager1989updating}. The existence of these techniques offers additional scalability, confirming that our framework is theoretically well-suited for massive, dynamically expanding systems.

\clearpage
\section{FLAN v2 Models and Datasets}
\label{sec:appendix_b}
We present the full list of $92$ FLAN v2 datasets.

\begin{enumerate}
    \item adversarial\_qa\_dbidaf\_based\_on
    \item adversarial\_qa\_dbert\_answer\_the\_following\_q
    \item adversarial\_qa\_dbidaf\_question\_context\_answer
    \item adversarial\_qa\_dbidaf\_tell\_what\_it\_is
    \item adversarial\_qa\_droberta\_tell\_what\_it\_is
    \item amazon\_polarity\_User\_recommend\_this\_product
    \item anli\_r1
    \item app\_reviews\_categorize\_rating\_using\_review
    \item bool\_q
    \item dbpedia\_14\_given\_a\_list\_of\_category\_what\_does\_the\_title\_belong\_to
    \item definite\_pronoun\_resolution
    \item dream\_baseline
    \item dream\_read\_the\_following\_conversation\_and\_answer\_the\_question
    \item drop
    \item duorc\_ParaphraseRC\_answer\_question
    \item duorc\_ParaphraseRC\_movie\_director
    \item duorc\_ParaphraseRC\_Youtubeing
    \item duorc\_SelfRC\_generate\_question\_by\_answer
    \item duorc\_SelfRC\_Youtubeing
    \item duorc\_SelfRC\_title\_generation
    \item gem\_e2e\_nlg
    \item gem\_web\_nlg\_en
    \item glue\_cola
    \item glue\_mrpc
    \item glue\_sst2
    \item glue\_wnli
    \item wiki\_hop\_original\_choose\_best\_object\_interrogative\_2
    \item imdb\_reviews\_plain\_text
    \item kilt\_tasks\_hotpotqa\_complex\_question
    \item lambada
    \item math\_dataset\_algebra\_linear\_1d
    \item sciq\_Multiple\_Choice\_Question\_First
    \item newsroom
    \item ropes\_prompt\_beginning
    \item qasc\_is\_correct\_1
    \item qasc\_is\_correct\_2
    \item qasc\_qa\_with\_combined\_facts\_1
    \item qasc\_qa\_with\_separated\_facts\_3
    \item qasc\_qa\_with\_separated\_facts\_5
    \item quac
    \item quail\_context\_description\_Youtube\_text
    \item quail\_context\_Youtube\_description\_id
    \item quail\_context\_Youtube\_description\_text
    \item quail\_context\_question\_description\_answer\_id
    \item quail\_context\_question\_description\_answer\_text
    \item quail\_description\_context\_Youtube\_id
    \item quail\_no\_prompt\_text
    \item quarel\_choose\_between
    \item quarel\_do\_not\_use
    \item quarel\_heres\_a\_story
    \item quarel\_logic\_test
    \item quarel\_testing\_students
    \item quartz\_having\_read\_above\_passage
    \item quartz\_read\_passage\_below\_choose
    \item quoref\_Find\_Answer
    \item quoref\_Found\_Context\_Online
    \item quoref\_Guess\_Title\_For\_Context
    \item race\_high\_Select\_the\_best\_answer
    \item race\_middle\_Is\_this\_the\_right\_answer
    \item race\_middle\_Select\_the\_best\_answer
    \item race\_middle\_Taking\_a\_test
    \item ropes\_prompt\_bottom\_hint\_beginning
    \item sciq\_Direct\_Question\_Closed\_Book\_
    \item social\_i\_qa\_Generate\_the\_question\_from\_the\_answer
    \item squad\_v1.1
    \item squad\_v2.0
    \item super\_glue\_wic
    \item super\_glue\_wsc.fixed
    \item trec
    \item true\_case
    \item web\_questions\_get\_the\_answer
    \item wiki\_bio\_comprehension
    \item wiki\_bio\_guess\_person
    \item wiki\_bio\_key\_content
    \item wiki\_bio\_who
    \item wiki\_hop\_original\_choose\_best\_object\_affirmative\_1
    \item wiki\_hop\_original\_choose\_best\_object\_interrogative\_1
    \item wiki\_hop\_original\_generate\_subject
    \item wiki\_qa\_automatic\_system
    \item wiki\_qa\_found\_on\_google
    \item wiki\_qa\_Is\_This\_True\_
    \item wiki\_qa\_Jeopardy\_style
    \item wiki\_qa\_Topic\_Prediction\_Answer\_Only
    \item wiqa\_effect\_with\_label\_answer
    \item wiqa\_what\_is\_the\_final\_step\_of\_the\_following\_process
    \item wiqa\_what\_might\_be\_the\_last\_step\_of\_the\_process
    \item wiqa\_what\_is\_the\_missing\_first\_step
    \item wmt16\_translate\_ro-en
    \item wiqa\_which\_of\_the\_following\_is\_the\_supposed\_perturbation
    \item wmt16\_translate\_tr-en
    \item word\_segment
    \item yelp\_polarity\_reviews
\end{enumerate}

\end{document}